\documentclass[a4paper, 10pt, conference]{ieeeconf} 
\IEEEoverridecommandlockouts       
\usepackage{amsmath,amssymb,amsfonts}
\usepackage{algorithm}
\usepackage{algpseudocode}
\usepackage{array}
\usepackage{multirow}
\usepackage{multicol}
\usepackage{makecell}
\usepackage{subcaption}
\usepackage{textcomp}
\usepackage{stfloats}
\usepackage{url}
\usepackage{booktabs}
\usepackage{verbatim}
\usepackage{graphicx}
\usepackage{mathrsfs,color}
\usepackage{caption}
\graphicspath{ {image/} }
\usepackage{bm}
\usepackage{tikz}
\usepackage{epstopdf}
\usepackage{cite}
\usepackage{arydshln}
\usepackage{hyperref}
\usepackage{bm}
\usepackage{nicematrix}

\newtheorem{theorem}{Theorem}

\newtheorem{lemma}{Lemma}

\newtheorem{assumption}{Assumption}

\newcommand\vn{\vec{\mathbf{n}}}

\newcommand\rc{{}^\mathcal{C}}
\newcommand\rw{{}^\mathcal{W}}

\newcommand\subcubeu{\boldsymbol{\mathcal{C}}_{\vec{\mathbf{u}}}}

\usepackage{booktabs}

\begin{document}
	\title{\LARGE \bf SCORE: Saturated Consensus Relocalization in Semantic Line Maps}
	\author{
		Haodong Jiang$^{\ast1}$, Xiang Zheng$^{\ast1}$, Yanglin Zhang$^{\ast1}$, Qingcheng Zeng$^{2}$, Yiqian Li$^{1}$, Ziyang Hong$^{1}$, Junfeng Wu$^{1}$
		\thanks{
			$^\ast$:Equal contribution.
			$^1$Haodong Jiang, Xiang Zheng, Yanglin Zhang, Yiqian Li, Ziyang Hong, and Junfeng Wu are with School of Data Science, The Chinese University of Hong Kong, Shenzhen, P. R. China, {\tt\small \{haodongjiang, 224045013, 119010446,yiqianli\}@link.cuhk.edu.cn, \{hongziyang,junfengwu\}@cuhk.edu.cn}. 
			$^2$ Qingcheng Zeng is with Robotics and Autonomous Systems Thrust, System Hub, The Hong Kong University of Science and Technology(Guangzhou), P. R. China, {\tt\small qzeng450@connect.hkust-gz.edu.cn}.
		}%
	}
	\maketitle
	\begin{abstract}
		This is the arXiv version of our paper published on IROS 2025. We present SCORE, a visual relocalization system that achieves unprecedented map compactness by adopting semantically labeled 3D line maps. SCORE requires only 0.01\%–0.1\% of the storage needed by structure-based or learning-based baselines, while maintaining practical accuracy and comparable runtime. The key innovation is a novel robust estimation mechanism, \textit{Saturated Consensus Maximization}~(Sat-CM), which generalizes classical \textit{Consensus Maximization}~(CM) by assigning diminishing weights to inlier associations according to maximum likelihood with probabilistic justification. Under extreme outlier ratios (up to 99.5\%) arising from one-to-many ambiguity in semantic matching, Sat-CM enables accurate estimation when CM fails. To ensure computational efficiency, we propose an accelerating framework for globally solving Sat-CM formulations and specialize it for the Perspective-n-Lines problem at the core of SCORE. 
	\end{abstract}
	\section{Introduction}\label{intro}
\begin{figure}[!t]
		\centering
		\includegraphics[width=\linewidth]{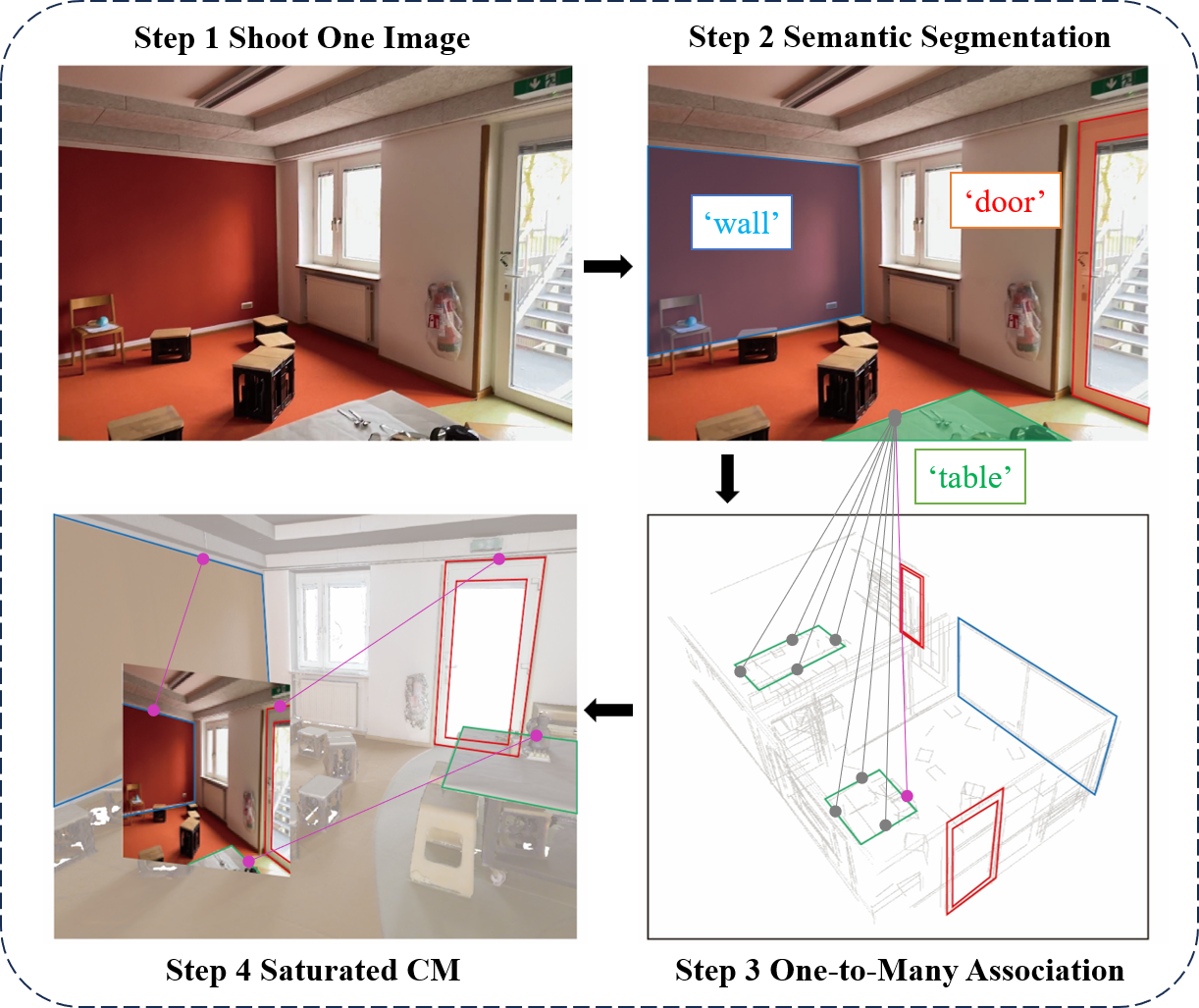}
		\caption{A kidnapped robot relocalizes itself by associating 2D lines in an image and 3D map lines based on semantic labels, and solving a perspective-n-lines problem. We propose \textbf{Saturated Consensus Maximization} to address the extremely high outlier ratio caused by one-to-many associations.}
		\label{fig::teaser}
	\end{figure}

	Visual relocalization refers to the task of estimating a camera's pose in a known environment from an input image—a critical capability for mobile robotics, augmented reality, and related applications. This process relies on stored scene representations, which vary across methods: some employ 3D map points with visual descriptors~\cite{sattler2015hyperpoints,tran2018device,camposeco2019hybrid,yang2022scenesqueezer,sarlin2019coarse}, others leverage deep feature maps from posed reference images~\cite{sarlin2021back,revaud2024sacreg}, or encode scene geometry into weights of neural networks~\cite{brachmann2017dsac,bui2024representing}. While these paradigms differ in how they represent and utilize scene information, they face a shared trade-off between representation compactness and estimation accuracy. In this work, we advance towards unprecedented compactness while maintaining practical accuracy with two ingredients as illustrated in Fig.~\ref{fig::teaser}: a semantically labeled 3D line map as scene representation, and \textit{Saturated Consensus Maximization}~(Sat-CM) as an enabling robust mechanism for resolving one-to-many ambiguous association. We adopt 3D lines to represent scene geometry due to their ubiquity in man-made environments and ability to capture structural elements more compactly than points. And we adopt semantics as the visual descriptor of map lines for three reasons. Firstly, semantics is a naturally compressed descriptor which takes only an integer label to store given the dictionary, consuming far less memory than fine-grained descriptors like SIFT~\cite{lowe2004distinctive} and SuperPoint~\cite{ detone2018superpoint}. Secondly, semantics exhibit strong invariance to partial occlusions and viewpoint changes, which are common challenges to line descriptors~\cite{pautrat2021sold2}. Lastly, The increasing availability of accurate segmentation models (e.g., SAM2~\cite{ravi2024sam}) and vision-language systems (e.g., GPT-4~\cite{achiam2023gpt}) makes semantic labeling increasingly practical. 
	
	However, this highly compact approach introduces a critical challenge: associating 2D-3D lines via semantic labels causes one-to-many ambiguity and extreme outlier ratios (up to 99.5\% in our experiments) for pose estimation. Although prior works have acknowledged similar ambiguity when using simple or quantized visual descriptors~\cite{sattler2015hyperpoints,tran2018device,camposeco2019hybrid}, the field still lacks a principled methodology for handling one-to-many association. Current approaches typically avoid rather than address association ambiguity - for instance, the widely-used ratio test~\cite{lowe2004distinctive} explicitly rejects ambiguous matches by requiring a dominant best candidate. In contrast, we embrace ambiguous associations through a principled robust mechanism, Sat-CM, which generalizes the classical \textit{consensus maximization}~(CM) method by assigning a diminishing weight for inlier associations according to likelihood. To integrate Sat-CM into our relocalization pipeline, named SCORE, we develop a general accelerated global search framework for Sat-CM problems, and devise a specialized solution for the perspective-n-lines~(PnL) problem. We summarize our \textbf{key contributions} as follows:
	\begin{enumerate}
		\item \textbf{Ultra-Compact Visual Relocalization:} we push the boundaries of map compactness for visual relocalization by adopting a semantically labeled 3D line map. 
		\item \textbf{Novel Robust Mechanism:} to address one-to-many ambiguity inherent in semantic association, we propose the Sat-CM method which generalizes classic CM method and evaluates inlier associations according to the maximum likelihood criteria.
		\item \textbf{Accelerated Global Solver:} we develop an accelerated global search framework for general Sat-CM problems, and apply it to solve the PnL problem central to our pipeline based on rigorous interval analysis.
		\item \textbf{Comprehensive Evaluation:} we demonstrate superiority of Sat-CM over CM, and evaluate practicality of SCORE with extensive experiments on the ScanNet++~\cite{yeshwanthliu2023scannetpp} dataset. Enabled with the power of Sat-CM and an accelerated global search algorithm, SCORE achieves practical accuracy within comparable runtime, while consuming only 0.01\% to 0.1\% storage of representative baselines. 
	\end{enumerate}
	\section{Related Work}
	\subsection{Visual Relocalization and Storage Burden}
	\textbf{Structure-based} visual relocalization methods establish explicit 2D-3D associations through detected local point features~\cite{lowe2004distinctive,detone2018superpoint} and their matches~\cite{lindenberger2023lightglue}, often aided by image retrieval~(IR) techniques~\cite{arandjelovic2016netvlad}. Recent advances~\cite{liu20233d,pautrat2023gluestick,liu2024lightweight} incorporate line segments to exploit structural regularities in man-made environments. While accurate, these methods maintain memory-intensive 3D maps with heavy visual descriptors (128 bytes per SIFT or 1 KB per SuperPoint descriptor vs. 12 bytes per 3D coordinate). \textbf{End-to-end learning methods} bypass explicit association by encoding scenes in neural network weights~\cite{brachmann2017dsac,bui2024representing}. Though elegant, these approaches are inherently scene-specific, requiring adaptation for new environments. Hybrid solutions~\cite{sarlin2021back,revaud2024sacreg} mitigate this issue by regressing poses relative to retrieved reference images with pre-established 2D-3D associations. However, they still demand substantial storage for either the reference images or their deep feature maps, which consume hundreds of KB per image even after quantization~\cite{revaud2024sacreg}.
	
	\subsection{Compact Representation and One-to-Many Ambiguity}
	To relieve the burden of storing memory-intensive visual descriptors, previous works propose to keep an informative subset of the 3D quantities~\cite{mera2020efficient}, quantize the visual descriptors~\cite{sattler2015hyperpoints, tran2018device}, or adopt a hybrid approach~\cite{camposeco2019hybrid, yang2022scenesqueezer} to reduce the map size. The one-to-many ambiguity in association arises along with descriptor quantization, while existing solutions are fundamentally limited. Sattler et al.~\cite{sattler2015hyperpoints} prune non-unique associations using co-visibility graphs, but risk prematurely discarding real associations. Recent RANSAC variants~\cite{tran2018device,camposeco2019hybrid} generate hypotheses exclusively from sampling unambiguous correspondences and utilize ambiguous associations only for evaluating the pose hypotheses. These variants fail when unique associations are unavailable and remain constrained by a CM formulation, which is not suitable under one-to-many ambiguity.  
	
	\section{Preliminary}
	\begin{figure}[!tbp]
		\centering
		\includegraphics[width=0.8\linewidth]{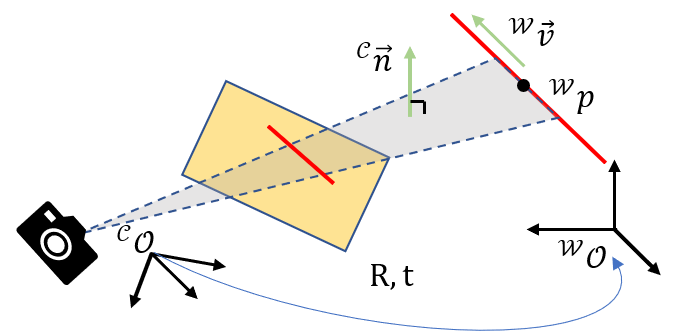}
		\caption{Projection model underlying the PnL problem.}
		\label{fig::line_projection}
	\end{figure}
	\textbf{Notations:} we use the notation $\mathbf{a}\bullet\mathbf{b}:= \mathbf{a}^\top\mathbf{b}$, and use $(\mathbf{a},\mathbf{b}):=\begin{bmatrix}
		\mathbf{a} \\ \mathbf{b}
	\end{bmatrix}$ to denote the concatenation of two vectors. We use the notation ${}^{\mathcal{F}}\mathbf{a}$ to highlight that $\mathbf{a}$ is observed in the reference frame ${\{\mathcal{F}}\}$. Specifically, we denote the normalized camera frame as ${\mathcal{C}}$, and the world frame as ${\mathcal{W}}$. We use $\mathbf{1}\{\cdot\}$ to denote the indicator function, which equals one if the event inside the bracket happens and equals zero otherwise. 
	
	Consider a 2D line in the image which writes as follows in the pixel coordinate: 
	$
	\begin{bmatrix}
		A&B&C
	\end{bmatrix}(u,v,1) = 0.
	$
	Given the camera intrinsic matrix $\mathbf{K}$, we can write it in the normalized image coordinate as
	$
	\begin{bmatrix}
		A_c&B_c&C_c
	\end{bmatrix}(x,y,1)=0,
	$
	with $\begin{bmatrix}
		A_c&B_c&C_c
	\end{bmatrix}=\begin{bmatrix}
		A&B&C
	\end{bmatrix}\mathbf{K}$. We use the normalized coefficient vector $\rc\vn$ to parameterize a 2D line $l$ in $\{\mathcal{C}\}$:
	$$
	\rc\vn=\frac{(A_c,B_c,C_c)}{\|(A_c,B_c,C_c)\|},~~l:=\{\rc\mathbf{p}\in\mathbb{R}^2|\rc\vn\bullet(\rc\mathbf{p},1)=0\}.
	$$
	We refer to $\rc\vn$ as the normal vector since it is perpendicular to the plane passing through the camera origin and $l$. As for a 3D line $L$ observed in the world coordinate, we parameterize it with a point $\rw\mathbf{p}_0\in\mathbb{R}^3$ and a direction vector $\rw\vec{\mathbf{v}}\in\mathbb{S}^2$:
	$$
	L:=\{\rw\mathbf{p}\in\mathbb{R}^3|(\rw\mathbf{p}-\rw\mathbf{p}_0)\times\rw\vec{\mathbf{v}}=0\}.
	$$
	Assume the relative transformation from the normalized camera frame to the world coordinate writes as follows
	$$
	{}^{\mathcal{W}}_{\mathcal{C}}\mathbf{T}=\begin{bmatrix}
		\mathbf{R} & \mathbf{t}\\\mathbf{0}^\top & 1
	\end{bmatrix}.
	$$
	Assume a 2D line $l$ in the image is the projection of a 3D line $L$ in the world map, the following two equations~\cite{liu1990determination} uniquely determine the projection:
	\begin{align}
		(\mathbf{R}\rc\vec{\mathbf{n}}) \bullet \rw\vec{\mathbf{v}} &= 0, \label{eqn::rotation_equation}\\
		(\mathbf{R}\rc\vec{\mathbf{n}})\bullet(\rw\mathbf{p}_0-\mathbf{t})&= 0.\label{eqn::translation_equation}
	\end{align}
	
	\section{Sat-CM Exemplified by the Robust PnL Problem with Semantic Matching}
	We address a challenging scenario where a kidnapped robot relocalizes itself with one image and an aggressively compressed scene map, which consists of 3D lines parameterized by their endpoints and semantic labels. The robot extracts 2D line segments from the image with semantic labels, and associates with map lines based on semantics. Without any other information, the robot associates each 2D line with all map lines of the same label, leading to a significant outlier ratio. While the CM method~\cite{fischler1981random} traditionally underpins robust estimation in robotics, its power is restricted under such one-to-many ambiguity. We therefore propose the Sat-CM method, which introduces a saturation function to handle more than one inlier associations. In this section, we formalize Sat-CM and motivate it with the PnL problem, where the robot pose is estimated using matched 2D image lines and 3D map lines.  
	
	\subsection{Saturated Consensus Maximization}
	Suppose we are given a sample set $\{\mathbf{x}_k\}$, a data set $\{\mathbf{y}_m\}$, and a residual function $f(\boldsymbol{\Theta}|\mathbf{x},\mathbf{y})$ w.r.t an unknown parameter $\boldsymbol{\Theta}$. Suppose under the ground truth $\boldsymbol{\Theta}^o$, the residual of a real sample-data association satisfies
	$
	|f(\boldsymbol{\Theta}^o|\mathbf{x},\mathbf{y})|\leq\epsilon,$ 
	where $\epsilon$ is a tolerance term to model against the effect of random noise. For each sample $\mathbf{x}_k$, there exists multiple matched data denoted as $\{\mathbf{y}_{m_k^{(i)}}\}$. The Sat-CM method estimates $\boldsymbol{\Theta}^o$ by solving the following problem:
	\begin{equation}\label{model::scm}
		\mathop{\rm max}\limits_{\boldsymbol{\Theta}} ~~\sum_{k=1}^K\sigma_k\left(\sum_{i} \boldsymbol{1}\{|f(\boldsymbol{\Theta}|\mathbf{x}_k,\mathbf{y}_{m_k^{(i)}})|\leq\epsilon\}\right),
	\end{equation}
	where $\sigma_k(N)=\sum_{n=1}^Nw_k(n)$ is referred to as the \textit{saturation function}, and one arrives at the classic CM formulation with $\sigma_k(N)=N$, $w_k(n)=1~\forall n\geq1$. For conciseness, we introduce several terminologies. Under a parameter hypothesis $\boldsymbol{\Theta}$, we refer to a sample-data pair $(\mathbf{x}_k,\mathbf{y}_{m_k^{(i)}})$ as an \textit{inlier association} given that $|f(\boldsymbol{\Theta}|\mathbf{x}_k,\mathbf{y}_{m_k^{(i)}})|\leq\epsilon$. We use the terminology \textit{settled} to describe the state of a sample $\mathbf{x}_k$ having at least one inlier association, and use \textit{unsettled} to describe the state of $\mathbf{x}_k$ having no inlier association. As the naming indicates, the saturation function assigns a diminishing weight $w_k(n)$ to each additional inlier of $\mathbf{x}_k$, and the total assigned weight saturates to a finite value.
	
	\subsection{PnL Problem under One-to-many Association}
	In the PnL problem, we treat 2D lines $l_k$ extracted from the image as samples $\mathbf{x}_k$, and the 3D map lines $L_m$ as data $\mathbf{y}_m$. We denote a 2D line with semantic label $s_k$, and a 3D line with semantic label $s_m$ respectively as:
	$$
	l_k: (\rc\vn_k,s_k),~~L_m: (\rw\mathbf{p}_m,\rw\mathbf{v}_m, s_m),
	$$  
	A semantic label $s$ is an integer id corresponding to a word in the dictionary, e.g., \textit{(1,`chair')} and \textit{(2,`table')}. For simplicity of notation and implementation, we treat a line with multiple labels as separate lines sharing the same geometric position but with different labels. We match each image line $l_k$ with all map lines of the same label, and denote this set of associations as $\{L_{m_k^{(i)}}\}$, with $\#\{L_{m_k^{(i)}}\}:=M_k$.
	
	The common practice in the robust PnL problem adopts the CM method and estimates rotation based on constraint~\eqref{eqn::rotation_equation} first~\cite{xu2016pose}. Under our setting, this problem writes:
	\begin{equation}\label{model::plain_rotation_CM} 
		\mathop{\rm max}\limits_{\mathbf{R}\in SO(3)} ~~\sum_{k=1}^K\sum_{i=1}^{M_k} \boldsymbol{1}\{|(\mathbf{R}\mathbf{}\rc\vec{\mathbf{n}}_k)\bullet\rw\vec{\mathbf{v}}_{m_k^{(i)}}|\leq\epsilon_r\},
	\end{equation}
	where the outer summation is over all 2D lines $l_k$, and the inner summation is over the associated 3D lines $L_{m_k^{(i)}}$. As illustrated by the following toy example, CM may flavor an unreasonable estimate under one-to-many ambiguity. Suppose the robot captures an image with 10 lines detected, and two rotation hypotheses $\mathbf{R}_1$ and $\mathbf{R}_2$ are evaluated by~\eqref{model::plain_rotation_CM}. Under hypothesis $\mathbf{R}_1$, lines $l_1$ to $l_{9}$ each has one inlier while line $l_{10}$ has no inlier, resulting a value of 9. Under hypothesis $\mathbf{R}_2$, lines $l_1$ to $l_{9}$ have no inlier, while line $l_{10}$ has 10 inliers, yielding a value of 10. Although $\mathbf{R}_2$ achieves a higher value in~\eqref{model::plain_rotation_CM}, one would intuitively consider $\mathbf{R}_1$ to be the better hypothesis since more 2D lines are settled under $\mathbf{R}_1$, indicating that a more diverse set of constraints—arising from the projections of these lines—are possibly satisfied. In case of $\mathbf{R}_2$, while $l_{10}$ yields 10 inliers, at most one of them is valid, and the inliers from fake associations do not contribute valid constraints on the rotation.  
	
	In pursuit of a systematic solution to the above issue, we adopt Sat-CM~\eqref{model::scm} and formulate the following problem: 
	\begin{equation}\label{model::saturated_rotation_CM}
		\mathop{\rm max}\limits_{\mathbf{R}\in SO(3)} ~~\sum_{k=1}^K\sigma_k\left(\sum_{i=1}^{M_k} \boldsymbol{1}\{ |\mathbf{}(\mathbf{R}\mathbf{}\rc\vec{\mathbf{n}}_k)\bullet\rw\vec{\mathbf{v}}_{m_k^{(i)}}|\leq \epsilon_r \}\right).
	\end{equation}
	The intuition from the previous toy example---that settling more 2D lines is better---may lead one to adopt the following `truncated' saturation function, which essentially counts the number of settled 2D lines: let $\sigma_k(N)=\sum_{n=1}^Nw_k(n)$,
	\begin{equation}\label{eqn::sat_func_truncate}
		\sigma_k(N)=\mathbf{1}\{N\geq1\},~w_k(n)=1\{n=1\}. 
	\end{equation}
	However, it performs poorly in highly ambiguous scenarios where bad hypotheses may tie~(or even beat) the good ones in the number of settled 2D lines. This occurs because the intuition underlying~\eqref{eqn::sat_func_truncate} fails to account for other important factors such as the number of associations $M_k$ for each 2D line $l_k$. For example, settling 2 lines each with 1 inlier out of 1000 putative associations is less likely to contribute to an accurate orientation than settling a single line with 1 inlier out of 2 putative associations. This underscores the necessity of principled saturation function design methodology.
	\subsection{Likelihood-Based Saturation Function Design}
	From a maximum likelihood standpoint, we propose a saturation function as follows: let $\sigma_k(N)=\sum_{n=1}^Nw_k(n)$:
	\begin{equation}\label{eqn::likelihood_sat_func}
		\sigma_k(N)=\log(1+C\frac{N}{M_k}),~w_k(n)=\log(\frac{M_k+nC}{M_k+(n-1)C}),
	\end{equation}
	where $C$ is a scaling constant to be derived in likelihood later. For each 2D line $l_k$, we introduce a hidden variable $\iota_k$ to indicate index of the real association: $\iota_k=i$ for association $(l_k,L_{m_k^{(i)}})$ to be real, while $\iota_k=0$ for all associations to be fake. Denote residual resulting from a hypothesis $\mathbf{R}$ and association $(l_k,L_{m_k^{(i)}})$ as $r_{k,i}(\mathbf{R}):=|(\mathbf{R}\rc\vec{\mathbf{n}_k})\bullet\rw\vec{\mathbf{v}}_{m_k^{(i)}}|$. Our derivations are based on two assumptions:
	\begin{assumption}\label{assump::associatoin}
		For each 2D line $l_k$, the association set $\{(l_k,L_{m_k^{(i)}})\}$ contains the real association with probability $q>0$, and each association is equally likely to be real, i.e.,
		$$
		\begin{cases}
			Pr(\iota_k=0)=1-q,\\
			Pr(\iota_k=i)=q/M_k,~~1\leq i\leq M_k.
		\end{cases}
		$$
	\end{assumption}
	\begin{assumption}\label{assump::uniform}
		With the ground truth rotation $\mathbf{R}^o$,
		$$
		\begin{cases}
			r_{k,i}(\mathbf{R}^o)\sim U(0,\epsilon_r),~\text{ conditioned on }\iota_k=i\\
			r_{k,i}(\mathbf{R}^o)\sim U(0,u_r),~\text{ conditioned on }\iota_k\neq i. 
		\end{cases}
		$$
		where $\epsilon_r$ is the same tolerance used in~\eqref{model::saturated_rotation_CM}, $u_r$ is the residual upper bound, and $u_r=1$ in the PnL rotation problem.
	\end{assumption}
	Note that Assumption~\ref{assump::associatoin} applies when the associations have equal confidence level, and Assumption~\ref{assump::uniform} extends the probabilistic justification of CM~\cite{antonante2021outlier}. We calculate the likelihood for line $l_k$ under a hypothesis ${\mathbf{R}}$ by marginalizing the hidden variable $\iota_k$ based on the above two assumptions. Denote the index set of the $N_k$ inliers under ${\mathbf{R}}$ as $\mathcal N_k$, with $N_k=\#\mathcal N_k$, and the vector stacking all residuals $r_{k,i}({\mathbf{R}})$'s as ${\mathbf{r}}_k$, where $i=1,\cdots,M_k$. Then, the likelihood for line $l_k$ writes
	\begin{align*}
		l_{{\mathbf{R}}}({\mathbf{r}}_k)
		=&p(\mathbf{{r}_k},\iota_k=0)+\sum_{\iota_k\in\mathcal N_k}p(\mathbf{{r}_k},\iota_k)+\sum_{\iota_k\not\in\mathcal N_k}p(\mathbf{{r}_k},\iota_k),\\
		=&(1-q)u_r^{-M_k}+N_kq/M_ku_r^{-(M_k-1)}\epsilon_r^{-1}+0\\
		=&u_r^{-M_k}(1-q)(1+CN_k/M_k),
	\end{align*}
	where $C:=u_r\epsilon_r^{-1}q(1-q)^{-1}$. We arrive at~\eqref{eqn::likelihood_sat_func} by taking logarithm and subtracting a common constant~(given $u_r$=1) in the likelihood of different $l_k$: $\log(1-q)-M_k\log(u_r)$.
	
\begin{figure}
		\centering
		\includegraphics[width=\linewidth]{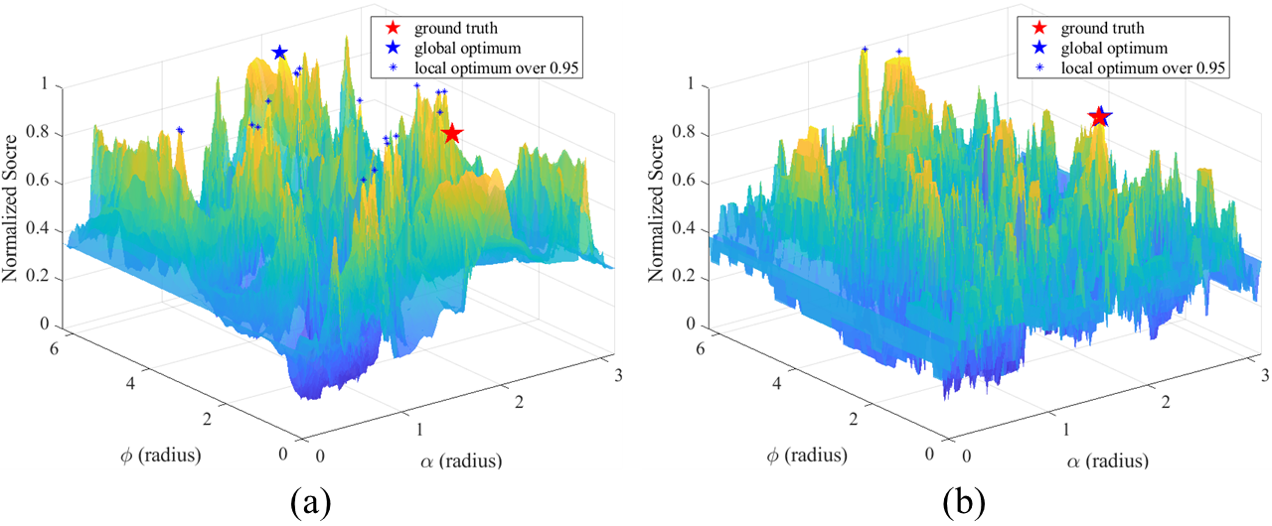}
		\caption{The normalized global landscapes under~(a) the CM problem~\eqref{model::plain_rotation_CM} and~(b) 
			the Sat-CM problem~\eqref{model::saturated_rotation_CM}. Each point is evaluated with a fixed rotation axis under polar coordinates $(\phi,\alpha)$ and the ad-hoc optimal rotation amplitude. We choose~\eqref{eqn::likelihood_sat_func} as the saturation function.} 
		\label{fig::sat_compare}
	\end{figure}

	We highlight two key observations on~\eqref{eqn::likelihood_sat_func}. \textbf{First,} the weight $w_k(n)$ assigned to the $n$th inlier association decreases w.r.t the total association number $M_k$, essentially penalizing the level of ambiguity. \textbf{Second,}~\eqref{eqn::likelihood_sat_func} assigns a diminishing weight $w_k(n)$ to  inliers, and the total assignment saturates to $\log(1+C)$ despite of $M_k$. This mechanism ensures that additional inliers contribute less to the overall consensus score, and prioritizes settling more lines over inflating the number of insignificant inliers. Although the diminishing rate of $w_k(n)$ is controlled by a hyperparameter $q$, estimators based on~\eqref{eqn::likelihood_sat_func} are not sensitive to the choice of $q$ in an appropriate range, as supported by our relocalization experiments. To enclose our discussion on function design, we present the global landscapes of the CM rotation problem~\eqref{model::plain_rotation_CM} and the Sat-CM one~\eqref{model::saturated_rotation_CM} in Fig.~\ref{fig::sat_compare}, where the 2D/3D associations come from one of the query image in our experiments. As indicated by Fig.~\ref{fig::sat_compare}, adopting Sat-CM with a likelihood-justified function design structurally reinforce the true value as the global optimum while suppressing ambiguous solutions. 
	
	Finally, given the global optimum $\hat{\mathbf{R}}^*_{\sigma}$ of~\eqref{model::saturated_rotation_CM} and the corresponding inliers, we further formulate the translation estimation problem using Sat-CM:
	\begin{equation}\label{model::satCM_trans}
		\mathop{\rm max}\limits_{\mathbf{t}\in\mathbb{R}^3}\sum_{k=1}^K\sigma_k\left(\sum_{i=1}^{M_k} \boldsymbol{1}\left\{ |\rw\vec{\mathbf{n}}_k^*\bullet(\rw\mathbf{p}_{m_k^{(i)}}-\mathbf{t})|\leq\epsilon_t \right\}\right),
	\end{equation}
	where $\rw\vec{\mathbf{n}}_{k}^*:=\hat{\mathbf{R}}^*_{\sigma}\rc\vec{\mathbf{n}}_k$ is the rotated normal vector\footnote{Note that since $\rw\vec{\mathbf{n}}_{k}$ is not perfectly orthogonal to the direction vector $\rw\vec{\mathbf{v}}_{k_j}$, choosing different point $\rw\mathbf{p}_{k_j}$ on a map line influences the residual. In order to make problem~\eqref{model::satCM_trans} invariant with the choice of parameterizing point, we project $\rw\vec{\mathbf{n}}_{i}^*$ onto the null space of $\rw\vec{\mathbf{v}}_{k_j}$ in implementation.}.
	\section{Solving Sat-CM Efficiently with Dimension-Reduced Branch-and-Bound}\label{section::fast_solution_2_sat_cm}
	As an extension of the NP-hard CM problem~\cite{chin2018robust}, Sat-CM is also computationally challenging to solve. In this section, we propose an accelerated branch-and-bound~(BnB) algorithm for Sat-CM problems, which reduces the branching dimensions by one through novel bounding techniques as inspired by a recent accelerating framework~\cite{zhang2024accelerating}. Next, we deploy this general algorithm to solver the PnL problems~\eqref{model::saturated_rotation_CM} and \eqref{model::satCM_trans}. While a recent work~\cite{huang2024efficient} proposes a similar global rotation solver for PnL, our FGO-PnL~(Fast and Globally Optimal) solver provides three key advances. \textbf{Firstly}, FGO-PnL accommodates the more general Sat-CM formulation. \textbf{Secondly}, rigorous interval analysis stands behind FGO-PnL to guarantee bound validity, while~\cite{huang2024efficient} uses heuristic approximation in their implementation. \textbf{Lastly,} FGO-PnL also includes an accelerated global translation solver. We release in our Git-Hub repository\footnote{https://github.com/LIAS-CUHKSZ/SCORE} Matlab and parallelized C++ implementations for FGO-PnL, which provide modular saturation function interfaces for easy customization and backward compatibility with CM via choosing $\sigma(N)=N$.
	\subsection{Accelerated Branch-and-Bound for Sat-CM}
\begin{figure}[!tbp]
		\centering
		\begin{subfigure}[b]{0.225\textwidth}
			\includegraphics[width=\textwidth]{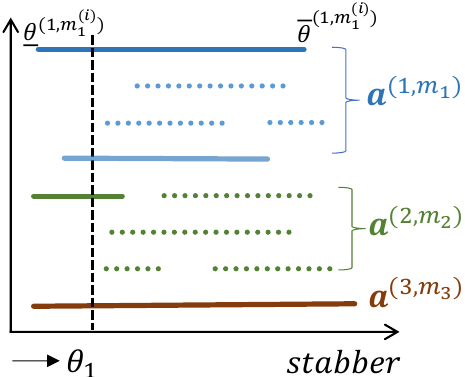}
			\caption{}
			\label{fig::SatIS_1}
		\end{subfigure}
		\begin{subfigure}[b]{0.215\textwidth}
			\includegraphics[width=\textwidth]{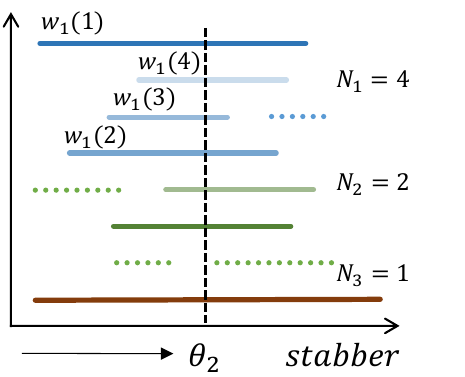}
			\caption{}
			\label{fig::SatIS_2}
		\end{subfigure}
		\caption{Saturated interval stabbing. The increased transparency of intervals represents adaptively decreased weights for each additional inlier. Algorithm~\ref{algorithm::sat_is} sweeps the range of stabber $\theta$ from left to right and record the highest value achieved along the way.~(a): $v(\theta_1)=\sigma_1(2)+\sigma_2(1)+\sigma_3(1)$,~(b):$v(\theta_2)=\sigma_1(4)+\sigma_2(2)+\sigma_3(1)$.}
		\label{fig::Sat_IS}
	\end{figure}

	For clarity, we use $\mathbf{a}^{(k,m_k^{(i)})}:=(\mathbf{x}_k,\mathbf{y}_{m_k^{(i)}})$ to denote an association. We start with a 1D Sat-CM problem:
	\begin{equation}\label{model::1-D_sat_CM}
		\max_{\theta}\sum_{k=1}^{K}\sigma_k\left(\sum_{i=1}^{M_k}\mathbf{1}\{|f(\theta|\mathbf{a}^{(k,m_k^{(i)})})|\leq\epsilon\}\right).
	\end{equation}
	Assume the residual function is continuous in the parameter $\theta$, we obtain by interval analysis:
	$$
	|f(\theta|\mathbf{a}^{(k,m_k^{(i)})})|\leq\epsilon \Leftrightarrow \theta\in [\underline{\theta}^{(k,m_k^{(i)})},\bar{\theta}^{(k,m_k^{(i)})}], 
	$$
	in which we assume a single interval is obtained for notational clarity. In practice, an association $\mathbf{a}^{(k,m_k^{(i)})}$ may induce multiple disjoint intervals, while it does not affect the proposed algorithm. Based on these intervals, we arrive at an equivalent problem for~\eqref{model::1-D_sat_CM}, which we refer to as saturated interval stabbing~(Sat-IS) and illustrate in Fig.~\ref{fig::Sat_IS}:
	\begin{equation}\label{model::1-D_sat_is}
		\max_{\theta}\sum_{k=1}^{K}\sigma_k\left(\sum_{i=1}^{M_k}\mathbf{1}\{\theta\in[\underline{\theta}^{(k,m_k^{(i)})},\bar{\theta}^{(k,m_k^{(i)})}]\}\right).
	\end{equation}
	Denote $M:=\sum M_k$, we can solve~\eqref{model::1-D_sat_is} with $O(M)$ space and $O(M\log{M})$ time using Algorithm~\ref{algorithm::sat_is}.
	\begin{algorithm}[!htbp]
		\caption{Saturated Interval Stabbing}
		\label{algorithm::sat_is}
		\begin{algorithmic}[1]
			\Statex \textbf{Inputs:} Intervals $\mathcal{I}_k=\{[\underline{\theta}^{(k,m_k^{(i)})},\bar{\theta}^{(k,m_k^{(i)})}]\}_{m_k^{(i)}}$ for $k=1,...,K$ and functions $\sigma_k(N)=\sum_{n=1}^N w_k(n)$.
			\Statex \textbf{Outputs:} optimal stabbers $\theta^*_{\{~\}}$, optimal value $v^*$.
			\State $I$ = sort endpoints in $\{\mathcal{I}_k\}_{k=1}^K$. 
			\State $v^*=0,~v=0,~N_k=0~(k=1,...,K)$.
			\For{$i=1$ to $len(I)$} 
			\If {$I(i)$ is a left endpoint from $\mathcal{I}_k$} 
			\State $N_k= N_k+1,~~v= v+w_k(N_k).$
			\If{$c>v^*$}
			\State $v^*= v,~~\theta^*_{\{~\}}= [I(i),I(i+1)].$
			\EndIf
			\Else 
			\State $v= v-w_k(N_k),~~N_k= N_k-1.$
			\EndIf
			\EndFor
			\State \Return{$\theta^*_{\{~\}}$, $v^*$}
		\end{algorithmic}
	\end{algorithm}
	
	As for a high-dimensional problem~\eqref{model::scm}, we distinguish one parameter $\theta$ with others: $\boldsymbol{\Theta}=(\boldsymbol{\Theta}_{:-1},\theta)$, and branches only over the space of $\boldsymbol{\Theta}_{:-1}$, as motivated by~\cite{zhang2024accelerating}. Denote the optimal value of~\eqref{model::scm} for $\boldsymbol{\Theta}_{:-1}$ constricted in a sub-cube $\boldsymbol{\mathcal{C}}$ and $\theta$ free as $r^*(\boldsymbol{\mathcal{C}})$. A \textbf{lower bound} for $r^*(\boldsymbol{\mathcal{C}})$ writes:
	\begin{equation}\label{eqn::lower_bound}
		\max_{\theta}\sum_{k=1}^{K}\sigma_k\left(\sum_{i=1}^{M_k}\mathbf{1}\{|f(\theta|\boldsymbol{\Theta}_{:-1}^{(c)}, \mathbf{a}^{(k,m_k^{(i)})})|\leq\epsilon\}\right),
	\end{equation}
	where $\boldsymbol{\Theta}_{:-1}^{(c)}$ is the center point of sub-cube $\boldsymbol{\mathcal{C}}$. Notice that the lower bound~\eqref{eqn::lower_bound} corresponds to a 1-D Sat-CM problem, and it can be efficiently solved by Algorithm~\ref{algorithm::sat_is}. As for the upper bound, we manage to find two bounding functions for each residual term:
	$$
	f_L^{(k,i)}(\theta)\leq f(\theta|\boldsymbol{\Theta}_{:-1},\mathbf{a}^{(k,m_k^{(i)})})\leq f_U^{(k,i)}(\theta),
	$$
	where the inequality holds for any $\boldsymbol{\Theta}_{:-1}\in\boldsymbol{\mathcal{C}}$. Given $f_L$ and $f_U$, we find an \textbf{upper bound} for $r^*(\boldsymbol{\mathcal{C}})$ as follows:
	\begin{equation}\label{eqn::upper_bound}
		\max_{\theta}\sum_{k=1}^{K}\sigma_k(
		\sum_{i=1}^{M_k}\mathbf{1}\{f_L^{(k,i)}(\theta)\leq\epsilon\textbf{ and } f_U^{(k,i)}(\theta)\geq-\epsilon\}).
	\end{equation}
	Similar to the lower bound, we obtain the upper bound~\eqref{eqn::upper_bound} by solving a Sat-IS problem with Algorithm~\ref{algorithm::sat_is}. Equipped with the above lower and upper bounds, we search the global optimum of a N-D Sat-CM problem by branching and bounding only a subspace of $N-1$ dimensions. For details in the BnB procedure, one can refer to our implementation and~\cite{zhang2024accelerating}. Next, we focus on finding the upper bounding functions $f_L$ and $f_U$ for the PnL problem. We put analysis corresponding to the translation part in Appendix~\ref{appendix::translation_bounding}, and focus on the challenging rotation problem. For conciseness, in the rest of this section, we refer to a general association as $\mathbf{a}:=(l,L)$, and denote the corresponding normal, direction vector and a 3D point on $L$ respectively as $\vec{\mathbf{n}}_a$, $\vec{\mathbf{v}}_a$ and $\mathbf{p}_a$. 
	\subsection{Find Bounding Functions for Rotation Problem in PnL}
	We parameterize rotation with a rotation axis $\vec{\mathbf{u}}\in\mathbb{S}^2$ and an amplitude $\theta\in[0,\pi]$. We choose $\theta$ as the distinguished parameter, and further parameterize $\vec{\mathbf{u}}$ by polar coordinates: 
	$$
	\vec{\mathbf{u}}=(\sin{\alpha}\cos{\phi}, \sin{\alpha}\sin{\phi},\cos{\alpha})~~ \alpha\in[0,\pi]~\phi\in[0,2\pi].
	$$
	We denote a sub-cube for axis $\vec{\mathbf{u}}$ in the polar coordinate as
	$$
	\subcubeu:=\{(\alpha,\phi)|\alpha\in[\underline{\alpha},\bar{\alpha}],\phi\in[\underline{\phi},\bar{\phi}]\},
	$$
	and denote its boundary as $\partial \subcubeu$. 
	
	Rewrite the residual function $f(\theta,\vec{\mathbf{u}}|\mathbf{a})$ for~\eqref{model::saturated_rotation_CM} as
	\begin{equation}\label{eqn::residual}
		\vec{\mathbf{n}}_a^\top  \vec{\mathbf{v}}_a +\sin{\theta}\vec{\mathbf{n}}_a^\top(\vec{\mathbf{u}}\times \vec{\mathbf{v}}_a)+(1-\cos{\theta})\vec{\mathbf{n}}_a^\top[\vec{\mathbf{u}}]_{\times}^2\vec{\mathbf{v}}_a,
	\end{equation}
	where $[\vec{\mathbf{u}}]_\times$ denotes the skew matrix of $\vec{\mathbf{u}}$. Notice that as a function of $\theta\in[0,\pi]$, the residual~\eqref{eqn::residual} is monotone w.r.t: 
	$$
	h_1(\vec{\mathbf{u}}|\mathbf{a}):=
	\vec{\mathbf{u}}^\top( \vec{\mathbf{v}}_a\times\vec{\mathbf{n}}_a),~~~~h_2(\vec{\mathbf{u}}|\mathbf{a}):=\vec{\mathbf{n}}_a^\top[\vec{\mathbf{u}}]_{\times}^2\vec{\mathbf{v}}_a.
	$$
	Based on monotonicity, we obtain the bounding functions for $f(\theta,\vec{\mathbf{u}}|\mathbf{a}),\vec{\mathbf{u}}\in\boldsymbol{\mathcal{C}}$ by finding the lower and upper values of spherical functions $h_1$ and $h_2$ for $\vec{\mathbf u}\in \boldsymbol{\mathcal{C}}_{\vec{\mathbf{u}}}$, denoted as $h_1^L$, $h_1^U$, $h_2^L$ and $h_2^U$. The bounding functions write:
	\begin{subequations}\label{eqn::bounding_function_for_rot}
		\begin{align}
			f_L(\theta)&= \vec{\mathbf{n}}_a^\top  \vec{\mathbf{v}}_a +h_1^{L}\sin{\theta}+h_2^L(1-\cos{\theta}),\\
			f_U(\theta)&= \vec{\mathbf{n}}_a^\top  \vec{\mathbf{v}}_a +h_1^{U}\sin{\theta}+h_2^U(1-\cos{\theta}).
		\end{align}
	\end{subequations}
	We summarize our results in the following two theorems. For conciseness, we first introduce several notations:
	$$
	\vec{\mathbf{m}}_a = \frac{\vec{\mathbf{v}}_a+\vec{\mathbf{n}}_a}{\|\vec{\mathbf{v}}_a+\vec{\mathbf{n}}_a\|},~~\vec{\mathbf{m}}_a^\perp=\frac{\vec{\mathbf{v}}_a-\vec{\mathbf{n}}_a}{\|\vec{\mathbf{v}}_a-\vec{\mathbf{n}}_a\|},~~\vec{\mathbf{c}}_a:=\frac{\vec{\mathbf{v}}_a\times\vec{\mathbf{n}}_a}{\|\vec{\mathbf{v}}_a\times\vec{\mathbf{n}}_a\|}.
	$$
	
	\begin{theorem}[Extreme Points for $h_2(\vec{\mathbf{u}}|\mathbf{a})$]\label{Theorem::h2}
		\quad
		\begin{enumerate}
			\item $\text{If} \pm\vec{\mathbf{m}}_a\in\boldsymbol{\mathcal{C}}_{\vec{\mathbf{u}}}, \arg\max_{\vec{\mathbf{u}}\in\subcubeu}h_2(\vec{\mathbf{u}}|\mathbf{a})=\pm\vec{\mathbf{m}}_a.$
			\item $\text{If} \pm\vec{\mathbf{m}}_a^\perp\in\subcubeu, \arg\min_{\vec{\mathbf{u}}\in\subcubeu}h_2(\vec{\mathbf{u}}|\mathbf{a})=\pm\vec{\mathbf{m}}_a^\perp.$
			\item $\text{Otherwise, the extreme points fall on } \partial \subcubeu.$
		\end{enumerate}
		
	\end{theorem}
	\begin{theorem}[Extreme Points for $h_1(\vec{\mathbf{u}}|\mathbf{a})$]\label{Theorem::h1}
		\quad
		\begin{enumerate}
			\item If $\vec{\mathbf{c}}_a\in\subcubeu$, $\arg\max_{\vec{\mathbf{u}}\in\subcubeu}h_1(\vec{\mathbf{u}}|\mathbf{a})=\vec{\mathbf{c}}_a.$ 
			\item If $-\vec{\mathbf{c}}_a\in\subcubeu$, $\arg\min_{\vec{\mathbf{u}}\in\subcubeu}h_1(\vec{\mathbf{u}}|\mathbf{a})=-\vec{\mathbf{c}}_a$. 
			\item Otherwise, the extreme points fall on $\partial \subcubeu$.
		\end{enumerate}
	\end{theorem}
	We present the proof for Theorem~\ref{Theorem::h2} in
	Appendix~\ref{appendix::h2} and omit the proof for Theorem~\ref{Theorem::h1} which is similar. For case 3) in Theorem~\ref{Theorem::h1}, we actually can explicitly write out the extreme points without the need to transverse $\partial\subcubeu$, and more explanation is provided in Appendix~\ref{appendix::h1}. 
	\section{Experiments on ScanNet++ Dataset}\label{section::experiment}
\begin{figure}
		\centering
		\includegraphics[width=0.99\linewidth]{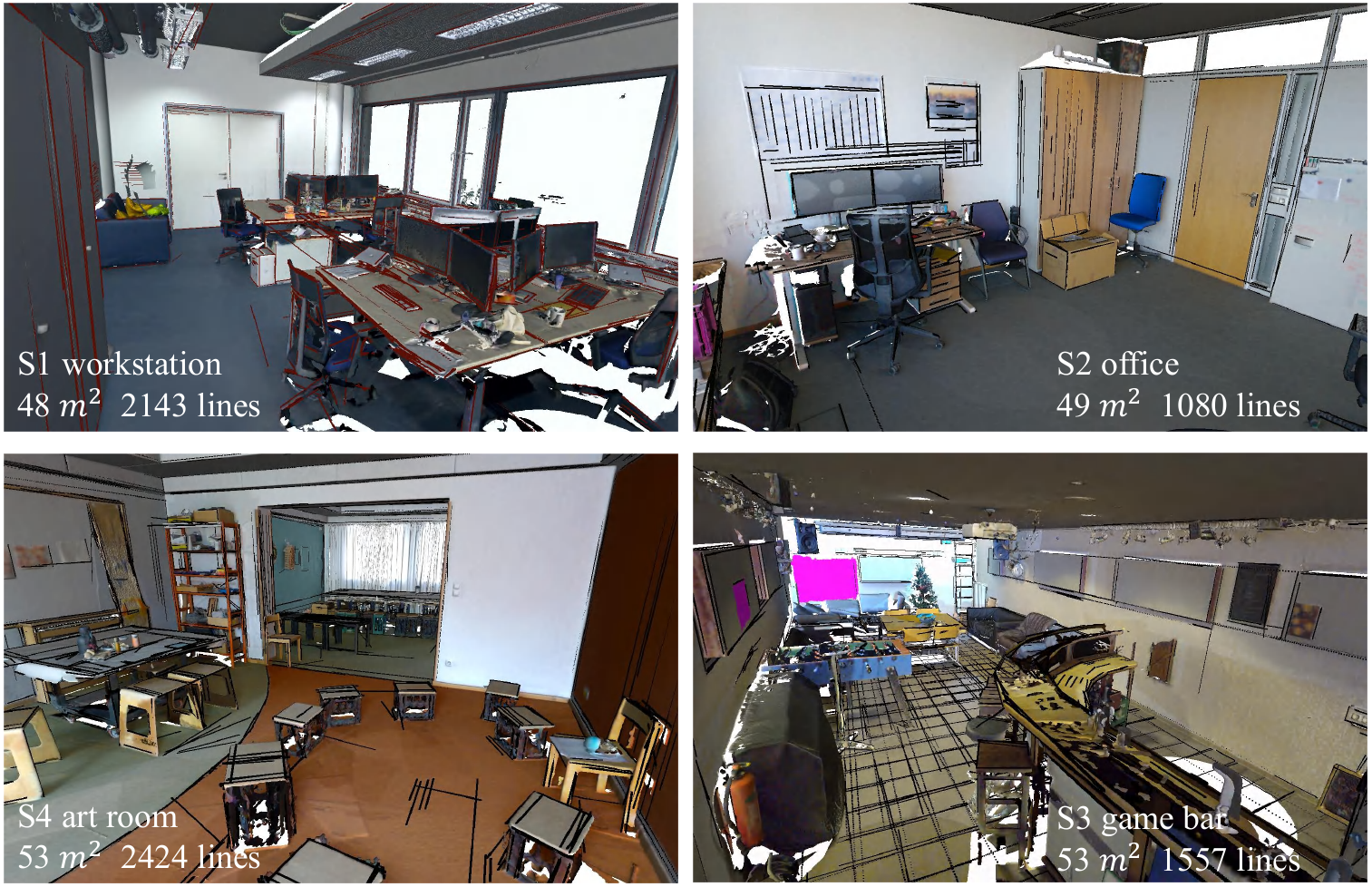}
		\caption{Meshes of the selected four scenes from ScanNet++ and overlapped with the constructed line map.}
		\label{fig::scene}
	\end{figure}

	We implement relocalization experiments on the ScanNet++ dataset~\cite{yeshwanthliu2023scannetpp} ,
	mainly for the convenience to construct a semantic line map. ScanNet++ provides for each scene an image sequence captured by an iPhone 13 Pro with default camera setting, and the user can render depth and semantic mask for each image from a human-annotated mesh. For each scene, we split the sequence of images into a reference set and a query set at a 7:1 ratio. We propose a pipeline to construct semantic line maps based on the reference images, and relocalize the query images in the constructed maps. Due to page limit, we present relevant content in Appendix~\ref{appendix::map_construction}. As shown in Fig.~\ref{fig::scene}, the quality of the constructed line maps is compromised when compared to the ones constructed by mature point-based pipelines like COLMAP~\cite{schoenberger2016sfm}, while, on the other hand, it is an ideal testbed for effectiveness of the Sat-CM mechanism and robustness of SCORE. 
	\subsection{Settings for Relocalization Experiments}
	We test SCORE with both ground-truth and predicted semantic labels for the query images. Our customized segmentation pipeline integrates RAM++\cite{huang2023open} for semantic tagging and Grounded-SAM\cite{ren2024grounded} for object detection: RAM++ tags images within the line map's dictionary, and Grounded-SAM segments objects based on the recognized words. And we use the released weights without fine-tuning for both models. In the line map, non-stationary object labels are excluded from the dictionary, such as \textit{bottle} and \textit{jacket}. While including these labels might improve relocalization accuracy, their transient nature in real-world environments renders such gains unrealistic. Additionally, we merge semantically ambiguous labels such as \textit{shelf} and \textit{bookshelf} to increase segmentation accuracy. These design choices underscore the critical role of dictionary engineering—a topic meriting deeper investigation in balancing accuracy with practical applicability.
	
	We evaluate our work from two perspectives. From the methodology perspective, we evaluate the proposed Sat-CM mechanism with different saturation functions and compare with the classic CM method. From the practicability perspective, we compare SCORE with baselines in terms of runtime, storage, and accuracy. We respectively choose hloc~\cite{sarlin2019coarse} and PixLoc~\cite{sarlin2021back} as representatives for the structure-based methods relying on high-dimensional visual descriptors, and the learning-based method based on deep image features. For fair comparison, all methods use the same splitting of reference and query images and the same image retrieval~(IR) results from NetVLAD~\cite{arandjelovic2016netvlad}, relocalizing query images within the sub-map observed by 12 retrieved images. We use consistent error thresholds for the FGO-PnL solver across settings: $\epsilon_r=0.015$ and $\epsilon_t=0.03$. We run all experiments on a PC equipped with an AMD Core 7950x CPU@4.5GHz, RAM@64GB and a GeForce RTX 5070Ti graphic card.
	
	\subsection{Sat-CM v.s. CM}
	We present here results of rotation estimation, which encounters the most severe ambiguity as the first step of a cascaded estimation procedure. In FGO-PnL, we utilize the IR results to effectively prune the space of rotation axis $\vec{\mathbf{u}}\in\mathbb{S}^2$ according to that of the first retrieved image. Specifically, we divide $\mathbb{S}^2$ into cubes with equal side length and restrict our search within the cube where the retrieved rotation axis resides in. The side length is chosen according to IR accuracy, and we present the results with side length $\pi$~(binary division) and $\pi/2$~(octal division), corresponding to a vague and moderate belief in the retrieval accuracy respectively. We use `PR' and `GT' in the legends to distinguish using predicted or ground truth semantic labels for the query images. As for the saturation functions, we use SCM0 in the legends for the naive truncated function~\eqref{eqn::sat_func_truncate}, and {\color{blue}SCM1} for the likelihood-based function~\eqref{eqn::likelihood_sat_func}. We choose $q=0.9$ for {\color{blue}SCM1}-GT and $q=0.5$ for {\color{blue}SCM1}-PR.  We record the maximal error when FGO-PnL returns multiple global optimum, while in the complete pipeline, subsequent translation estimation helps select from the rotation candidates. For comparison in translation estimation and sensitivity evaluation in parameter $q$, we refer the readers to Appendix~\ref{appendix::supplementray}.
	
	As evidenced by Table~\ref{table::rot_SCM_CM}, {\color{blue}SCM1} demonstrates consistent superiority across scenes and settings, and the performance lead is particularly pronounced under harsh conditions involving large search spaces and predicted semantic labels. While SCM0, which uses the naive truncation function~\eqref{eqn::sat_func_truncate}, yields lower accuracy than CM in most configurations, primarily due to susceptibility to multiple global optima. These results substantiate that Sat-CM, when incorporating a justified saturation function, enables accurate estimation under one-to-many ambiguity when CM fails. Scene S3~(game bar) exhibits the poorest results, which we attribute to compromised quality of the line map due to a cluttered layout, less accurate image pose and depth. We further observe that a more confined search space of rotation axis with side length $\pi/2$ enhances accuracy universally, since ambiguous candidates are pruned ahead of estimation. At last, we highlight the estimation difficulty when using predicted semantic labels, which can not be fully conveyed by the extremely high outlier ratio~(consistently over 99\%). As presented in Table~\ref{table::scene_info}, beyond erroneous predictions, missed detections also degrade performance by reducing usable 2D lines. 
	\begin{table}
		\centering
		\begin{tabular}{ccccc}
			\toprule
			&  S1    & S2     & S3    & S4\\
			\midrule
			\# map lines       &  2143  & 1080   & 1557  & 2424\\
			\# map points      & 164860 & 108726 &155673 & 185040\\
			wrong  label ratio & 17.2\% & 19.6\% & 19.4\% & 20.5\%\\
			missed label ratio & 27.0\% & 30.4\% & 30.7\% & 30.0\%\\
			\bottomrule    
		\end{tabular}
		\caption{Scene detailed information.}
		\label{table::scene_info}
	\end{table}
	
	\begin{table*}[htbp]
		\centering
		\caption{Compare CM with Sat-CM in rotation estimation.}
		\label{table::rot_SCM_CM}
		\begin{tabular}{ccccccccccccc} 
			\toprule
			\multirow{2}{*}{Method} & S1(workstation) &  S2(office) & S3(game bar) & S4(art room)           & S1 &  S2 & S3 & S4      & S1  &  S2 & S3 & S4 \\
			& \multicolumn{4}{c}{25\%/50\%/75\% error quantiles~($^\circ$)}         & \multicolumn{4}{c}{Recall at 5$^\circ$~(\%)} &\multicolumn{4}{c}{Median Outlier Ratio~(\%)}\\
			\midrule
			CM$^\pi$-GT       & 1.3/3.0/146.9 & 0.7/1.2/8.3 & 1.2/12.0/161.5 & 0.8/1.3/111.0 & 54 & 75 & 45 & 65 &  \multirow{3}{*}{95.7}&\multirow{3}{*}{94.7}&\multirow{3}{*}{94.9}&\multirow{3}{*}{95.7}\\ 
			SCM0$^\pi$-GT      & 1.6/3.5/179.3 & 1.6/2.7/179.7 & 2.5/8.0/179.9 & 2.2/4.9/179.7 & 56 & 58 & 41 & 51     &&&&                                                                                    \\
			{\color{blue}SCM1}$^\pi$-GT &\bf 0.9/1.4/3.1 &\bf 0.6/0.9/1.9 &\bf 0.8/1.7/4.4 &\bf 0.6/0.9/1.7 &\bf81 &\bf90 &\bf76 &\bf91 &&&&                                                                                    \\
			
			\midrule
			CM$^{\frac{\pi}{2}}$-GT       & 1.1/2.4/39.8 & 0.7/1.1/2.6 & 1.1/2.8/33.0 & 0.7/1.1/2.7 & 63 & 85 & 53 & 79 & \multirow{3}{*}{95.7}&\multirow{3}{*}{94.7}&\multirow{3}{*}{94.9}&\multirow{3}{*}{95.7}\\ 
			SCM0$^{\frac{\pi}{2}}$-GT      & 1.6/2.9/5.4 & 1.5/2.2/3.3 & 2.2/4.7/63.8 & 1.9/3.2/7.1 & 72 & 78 & 51 & 68  &&&&                                                                                    \\
			{\color{blue}SCM1}$^{\frac{\pi}{2}}$-GT &\bf0.8/1.3/2.5 &\bf0.6/0.9/1.5 &\bf0.8/1.6/3.7 &\bf0.6/0.9/1.5 &\bf90 &\bf93 &\bf80 &\bf95 &&&&                                                                                     \\
			
			\midrule
			CM$^\pi$-PR   &2.4/92.5/168.3 & 1.0/3.9/177.0 & 7.9/138.8/179.1 & 1.1/22.5/174.3 & 31 & 52 & 24 & 45&\multirow{3}{*}{99.4}&\multirow{3}{*}{99.5}&\multirow{3}{*}{99.3}&\multirow{3}{*}{99.4} \\
			SCM0$^\pi$-PR  &4.6/171.5/179.8 & 3.0/178.7/179.9 & 4.2/179.5/180.0 & 6.6/179.4/180.0 & 27 & 33 & 25 & 22 &&&&                                                                                      \\
			{\color{blue}SCM1}$^\pi$-PR &\bf1.6/2.6/90.5 &\bf0.7/1.6/10.8 &\bf1.1/3.7/149.7 &\bf0.8/1.8/72.0 &\bf64 &\bf73 &\bf56 &\bf69            &&&&                                                                               \\
			
			\midrule
			CM$^{\frac{\pi}{2}}$-PR   &2.0/28.9/53.7 & 0.9/2.0/18.2 & 2.6/26.7/61.2 & 1.0/2.8/41.2 & 39 & 68 & 37 & 55 &\multirow{3}{*}{99.4}&\multirow{3}{*}{99.5}&\multirow{3}{*}{99.3}&\multirow{3}{*}{99.4} \\
			SCM0$^{\frac{\pi}{2}}$-PR  &2.8/8.7/89.2 & 2.2/3.6/17.1 & 2.9/80.3/103.0 & 3.0/13.2/102.7 & 40 & 61 & 32 & 36   &&&&                                                                                      \\
			{\color{blue}SCM1}$^{\frac{\pi}{2}}$-PR &\bf1.4/2.3/5.3 &\bf0.7/1.4/2.8 &\bf1.0/2.4/4.6 &\bf0.7/1.3/3.6 &\bf74 &\bf82 &\bf76 &\bf80  &&&&                                                                                    \\
			
			\bottomrule
		\end{tabular}
	\end{table*}
	
	\subsection{SCORE v.s. Classic Relocalization Pipelines}
	We report the memory consumption for storing map geometry~(3D points or lines), cameras~(reference image poses and 2D-3D associations) and features~(visual descriptors or deep feature maps). We also report the median runtime which consists of feature extraction, feature association and estimation. For hloc~\cite{sarlin2019coarse}, we adopt SuperPoint~\cite{detone2018superpoint} for keypoint extraction and description~(stored in float 16), and use Lightglue~\cite{lindenberger2023lightglue} for keypoint association. For PixLoc~\cite{sarlin2021back}, we store the reference images and extract feature maps of the 12 retrieved images online. For SCORE, we adopt the truncated saturation function instead of the likelihood-based function in translation estimation. This is because we further utilize two physical constraints to prune the inliers obtained from solving~\eqref{model::satCM_trans}: the truly-associated 3D line resides in front of the camera and its projection intersects with the image. We choose the truncated saturation function in order to keep more translation candidates before pruning, and indeed observe a more robust performance~(presented in Appendix~\ref{appendix::supplementray}). As the above methods all rely on IR, we include the accuracy achieved by NetVLAD~\cite{arandjelovic2016netvlad} for reference. 
	
	As presented in Table~\ref{table::reloc_baselines}, SCORE achieves practical relocalization accuracy using the true semantic labels, with median errors below 8 cm and 1.5$^\circ$, though performance degrades when using predicted labels, particularly in translation. While PixLoc and hloc achieve finer-grained accuracy due to their dense point-based representations~(Table~\ref{table::scene_info}), SCORE provides radical memory efficiency (0.01\%–0.1\% storage overhead) while maintaining comparable estimation runtime thanks to an accelerated global solver and a parallelized C++ implementation. Compared with IR results achieved by NetVLAD, SCORE consistently improve over the median rotation error even when using predicted labels\footnote{Notice that the median rotation error reported in Table~\ref{table::reloc_baselines} is slightly better than those in Table~\ref{table::rot_SCM_CM}, since we use translation estimation to assist selection among multiple rotation candidates.}. Given true semantic labels, SCORE achieves better translation accuracy than IR except in S4 which has the most complete reference image database. To sum up, SCORE proves ultra-compact~(kB-scale) semantic line maps can sustain viable relocalization, while with semantic segmentation remaining the accuracy and runtime bottleneck.
	\begin{table*}
		\centering
		\caption{Baseline comparison in terms of runtime, memory consumption and accuracy.}
		\label{table::reloc_baselines}
		\begin{tabular}{cccccccccc} 
			\toprule
			\multirow{2}{*}{Method}        & median runtime            & S1 &  S2 & S3 & S4  &S1&  S2 & S3 & S4 \\
			& extract + match\&est & \multicolumn{4}{c}{storage: geometry\&cameras+feature}        & \multicolumn{4}{c}{Median Pose Error~($cm$/$^\circ$)}  \\
			\midrule
			NetVLAD~\cite{arandjelovic2016netvlad} & \bf 25+5 ms       & 0+7~MB    &0+5~MB    &0+6~MB    &0+11~MB   &7.8/4.5&  11.1/10.0 & 8.8/6.6 & 3.4/4.0  \\
			hloc~\cite{sarlin2019coarse}           & 30+260 ms       & 40+788~MB &38+611~MB &51+809~MB &61+951~MB &\bf 0.5/0.2&  0.7/0.2   &\bf 0.8/0.2    & \bf 0.5/0.1\\ 
			PixLoc~\cite{sarlin2021back}           & 1910+30 ms        & 40+132~MB &38+90~MB  &51+120~MB &61+192~MB &\bf 0.5/0.2&\bf 0.6/0.2   &\bf 0.8/0.2    &\bf 0.5/0.1\\
			\midrule
			SCORE$^\pi$-PR                       &1525+245 ms     &\multirow{4}{*}{\bf 158+17~KB}&\multirow{4}{*}{\bf 92+8~KB}&\multirow{4}{*}{\bf 107+12~KB}& \multirow{4}{*}{\bf 206+19~KB}   &13.3/2.4 & 7.7/1.5 & 100.7/2.5 & 21.2/1.7\\
			SCORE$^{\frac{\pi}{2}}$-PR           &1525+100 ms     &&&&                                                                                                          &10.6/2.2 & 6.8/1.2 & 32.3/2.3 & 14.4/1.2\\
			SCORE$^\pi$-GT                       & /+130 ms       &&&&                                                                                                          &5.9/1.4 & 3.8/0.8 & 7.1/1.5 & 5.0/0.9 \\
			SCORE$^{\frac{\pi}{2}}$-GT           & /+45 ms        &&&&                                                                                                          &5.9/1.3 & 3.8/0.8 & 6.9/1.4 & 4.9/0.8\\
			\bottomrule
		\end{tabular}
	\end{table*}
	
	\section{Conclusion and Future Works}
	We push the boundary of map compactness in visual relocalization through two key innovations: semantically labeled 3D line maps as an ultra-efficient scene representation, and Saturated Consensus Maximization as a foundational solution for one-to-many ambiguous associations. Experimental validation on ScanNet++ confirms Sat-CM’s robustness and demonstrates our pipeline’s practical viability despite extreme map compression. To transition toward real-world deployment, we identify three critical research directions. Firstly, develop fast and approximate solvers for Sat-CM to further reduce runtime. Secondly, explore low-dimensional descriptors combining semantic-level efficiency with improved accuracy and repeatability. Thirdly, incorporate both point and line features to offer a more resilient and fine-grained solution.
	\section*{Acknowledgment}
	We thank the reviewers for their valuable comments, Mingzhe Li for providing support for baseline implementation, Prof. Jianhua Huang and Prof. Li Jiang for discussion. 
	
	\bibliographystyle{IEEEtran}
	\bibliography{reference}

	\appendices
	\section{Proof for Theorem~\ref{Theorem::h2}}\label{appendix::h2}
	Recall that Theorem~\ref{Theorem::h2} characterizes  extreme points of the spherical function $h_2(\vec{\mathbf{u}}|\mathbf{a})$, with the rotation axis $\vec{\mathbf{u}}$ belonging to a sub-cube $\boldsymbol{\mathcal{C}}_{\vec{\mathbf{u}}}$ in polar coordinates. 
	
\begin{figure}[tbp]
		\centering
		\begin{subfigure}[b]{0.24\textwidth}
			\includegraphics[width=\textwidth]{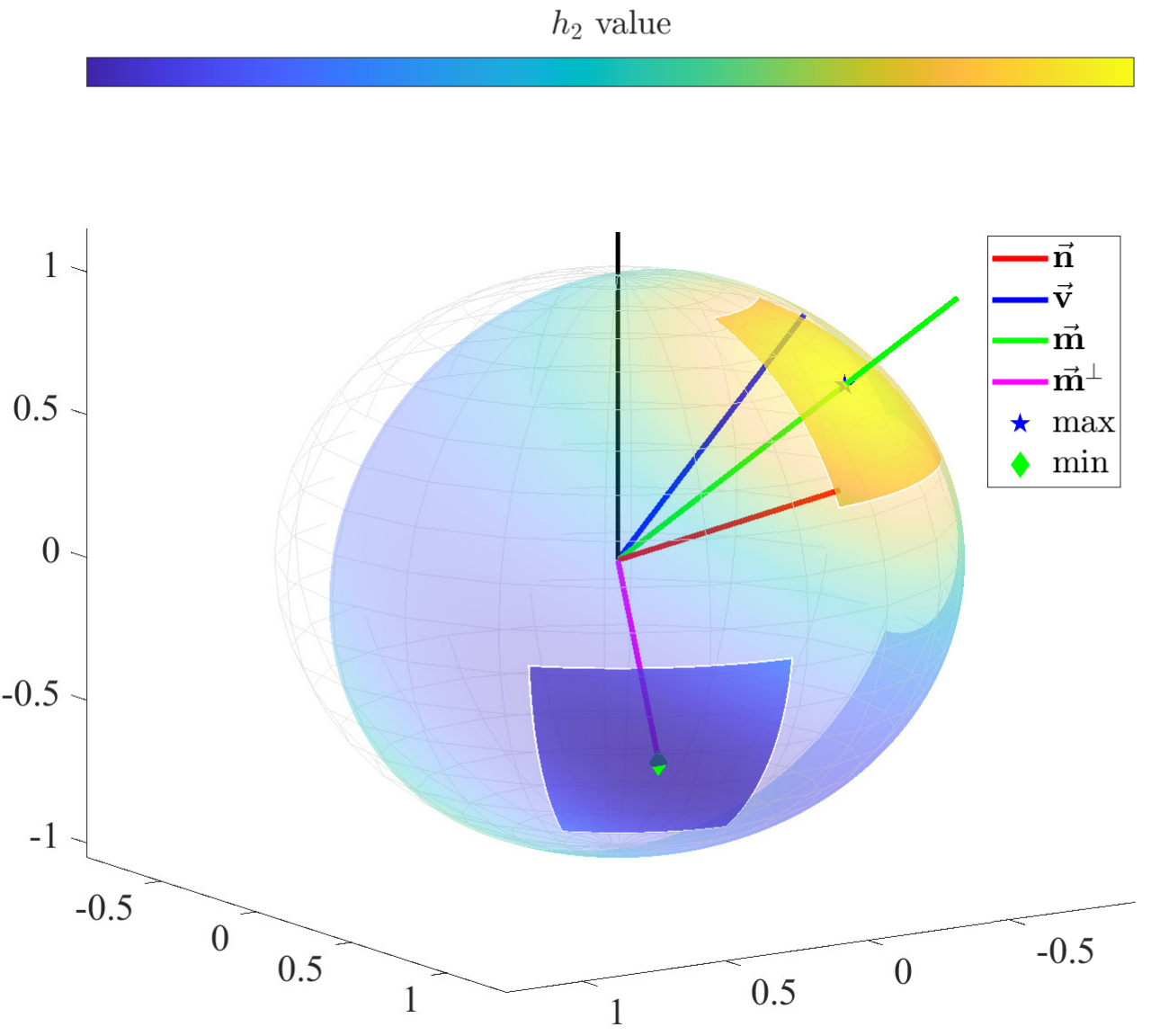}
			\caption{values over a half sphere}
			\label{fig::h2_half_sphere}
		\end{subfigure}
		\hfill
		\begin{subfigure}[b]{0.20\textwidth}
			\includegraphics[width=\textwidth]{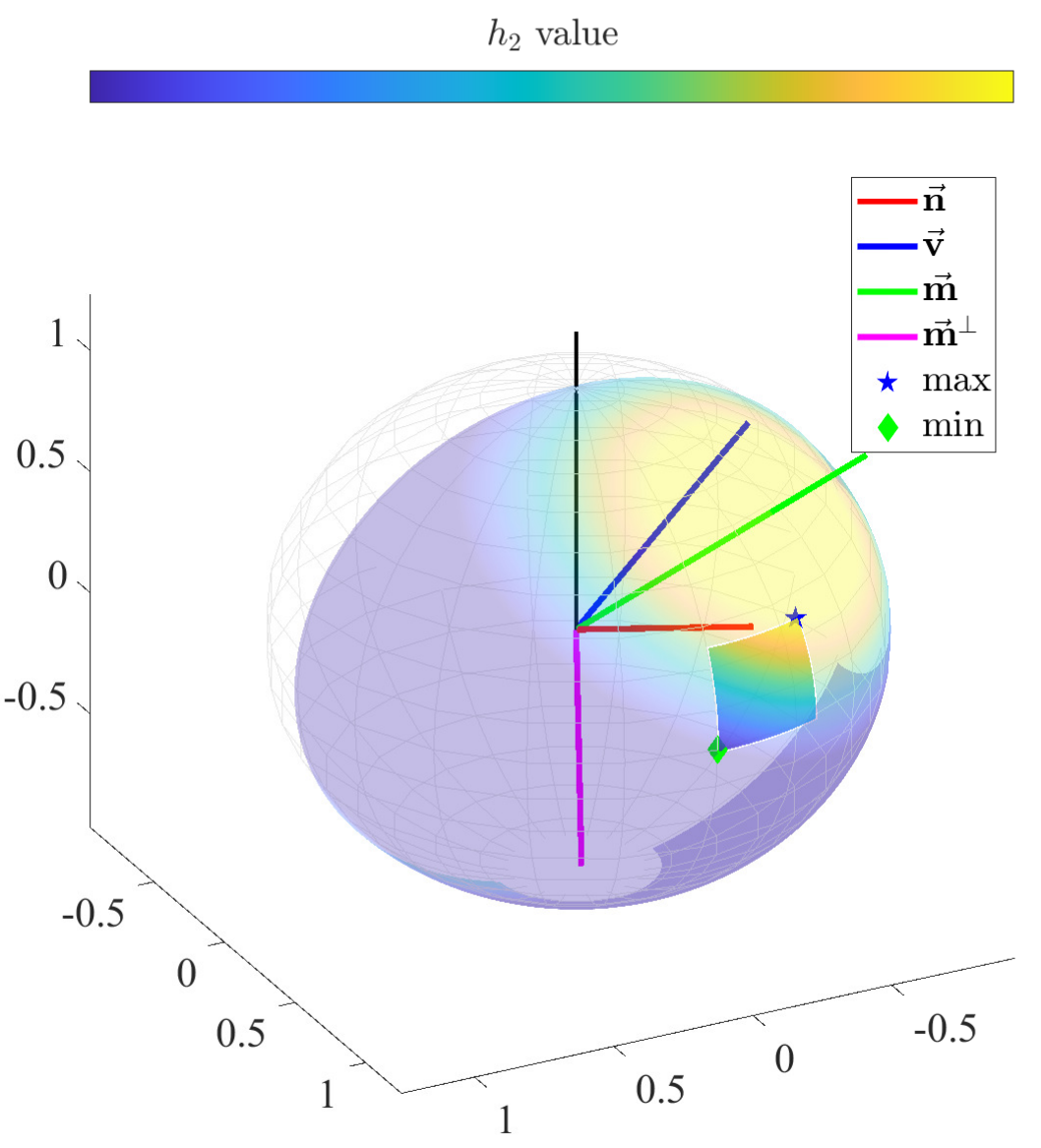}
			\caption{values over a sub-cube}
			\label{fig::h2_sub_cube}
		\end{subfigure}
		\caption{Landscapes of function $h_2(\vec{\mathbf{u}}|\mathbf{a})$.}
		\label{fig::h2_land_scape}
	\end{figure}

	\begin{proof}
		First, rewrite $h_2(\vec{\mathbf{u}}|\mathbf{a})$ as:
        \begin{align*}
            				h_2(\vec{\mathbf{u}}|\mathbf{a})
				&= \vec{\mathbf{n}}_a^\top[\vec{\mathbf{u}}]_{\times}^2\vec{\mathbf{v}}_a = \vec{\mathbf{n}}^\top_a(\vec{\mathbf{u}}\vec{\mathbf{u}}^\top - \mathbf{I}_3)\vec{\mathbf{v}}_a\\     &=\vec{\mathbf{u}}^\top(\vec{\mathbf{n}}_a\vec{\mathbf{v}}_a^\top)\vec{\mathbf{u}}-\vec{\mathbf{n}}_a^\top\vec{\mathbf{v}}_a\\       &=\vec{\mathbf{u}}^\top(\frac{\vec{\mathbf{n}}_a\vec{\mathbf{v}}_a^\top+\vec{\mathbf{v}}_a\vec{\mathbf{n}}_a^\top}{2})\vec{\mathbf{u}}-\vec{\mathbf{n}}_a^\top\vec{\mathbf{v}}_a.
        \end{align*}
		Next, we derive the critical points of $h_2(\vec{\mathbf{u}}|\mathbf{a})$ for $\vec{\mathbf{u}}\in\mathbb{S}^2$. Denote $\mathbf{M}_a:=\frac{\vec{\mathbf{n}}_a\vec{\mathbf{v}}_a^\top+\vec{\mathbf{v}}_a\vec{\mathbf{n}}_a^\top}{2}$, and take derivative of $h_2(\vec{\mathbf{u}}|\mathbf{a})$ w.r.t a perturbation $\boldsymbol{\delta}\in\mathbb{R}^3$ on the lie algebra as follows: 
		\begin{align*}
			&\frac{\partial h_2(\textbf{expm}([\boldsymbol{\delta}]_\times)\vec{\mathbf{u}}|\mathbf{a})}{\partial \boldsymbol{\delta}}\large|_{\boldsymbol{\delta}=0}\\
			&=\frac{\partial\left(\textbf{expm}([\boldsymbol{\delta}]_\times)\vec{\mathbf{u}}\right)^\top\mathbf{M}_a~\textbf{expm}([\boldsymbol{\delta}]_\times)\vec{\mathbf{u}}}{\partial \boldsymbol{\delta}}\large|_{\boldsymbol{\delta}=0},\\
			&=\frac{\partial \vec{\mathbf{u}}^\top(-[\boldsymbol{\delta}]_\times\mathbf{M}_a+\mathbf{M}_a[\boldsymbol{\delta}]_\times)\vec{\mathbf{u}}+o(\boldsymbol{\delta})}{\partial\boldsymbol{\delta}}\large|_{\boldsymbol{\delta}=0},\\
			&=2[\vec{\mathbf{u}}]_{\times}\mathbf{M}_a\vec{\mathbf{u}}=2\vec{\mathbf{u}}\times(\mathbf{M}_a\vec{\mathbf{u}}),
		\end{align*}
		where $\textbf{expm}(\cdot)$ is the matrix exponential operation, and we use the equality $[\mathbf{a}]_\times\mathbf{b}=-[\mathbf{b}]_\times\mathbf{a}$ in the last equation. To let the derivative equal zero, one has 
		$$
		\vec{\mathbf{u}}\parallel\mathbf{M}_a\vec{\mathbf{u}} \Leftrightarrow \vec{\mathbf{u}} \text{ is an eigenvector of }\mathbf{M}_a.
		$$
		Substitute and verify that $\vec{\mathbf{m}}_a$, $\vec{\mathbf{m}}_a^\perp$, and $\vec{\mathbf{c}}_a$ are three orthogonal eigenvectors of matrix $\mathbf{M}_a$, corresponding to a positive, negative and zero eigenvalue respectively. And $h_2(\vec{\mathbf{u}}|\mathbf{a})$ achieves the global maximum at $\pm\vec{\mathbf{m}}_a$, the global minimum at $\pm\vec{\mathbf{m}}_a^\perp$, and saddle points at $\pm\vec{\mathbf{c}}_a$. Given that $h_2(\vec{\mathbf{u}}|\mathbf{a})$ with $\vec{\mathbf{u}}\in\mathbb{S}^2$ has finite~(six) critical points, we conclude that its extreme values for $\vec{\mathbf{u}}\in\boldsymbol{\mathcal{C}}_{\vec{\mathbf{u}}}$ are achieved either at $\pm\vec{\mathbf{m}}_a$, $\pm\vec{\mathbf{m}}_a^\perp$, or at $\partial\boldsymbol{\mathcal{C}}_{\vec{\mathbf{u}}}$. 
	\end{proof}
	We visualize $h_2(\vec{\mathbf{u}}|\mathbf{a})$ on the sphere in Fig.~\ref{fig::h2_land_scape}.
	\section{Supplementary Results for Theorem~\ref{Theorem::h1}}\label{appendix::h1}
	Denote polar coordinates for $\vec{\mathbf{c}}_a:=\frac{\vec{\mathbf{v}}_a\times\vec{\mathbf{n}}_a}{\|\vec{\mathbf{v}}_a\times\vec{\mathbf{n}}_a\|}$ as $(\alpha_c,\phi_c)$:
    $$ h_1(\vec{\mathbf{u}}|\mathbf{a})=\sin{\alpha_c}\sin{\alpha}\cos{(\phi_c- \phi)} + \cos{\alpha_c}\cos{\alpha}+const.
    $$
The partial derivative of $h_1$ with respect to $\alpha$ and $\phi$ writes
	\begin{subequations}
		\begin{align}
			\frac{\partial h_1}{\partial \alpha}&=\sin{\alpha_c}\cos{\alpha}\cos{(\phi_c - \phi)}-\sin{\alpha}\cos{\alpha_c},\label{eqn::h1_partial_alpha}\\
			\frac{\partial h_1}{\partial \phi}&=\sin{\alpha_c}\sin{\alpha}\sin{(\phi_c - \phi)}.
			\label{eqn::h1_partial_phi}
		\end{align}
	\end{subequations}
	For conciseness, we focus on the case where both the sub-cube $\subcubeu$ and $\vec{\mathbf{c}}_a$ are in the east-hemisphere, i.e., $\phi\in[0,\pi]$ and $\phi_c\in[0,\pi]$. The following arguments can be easily adapted to the other cases. Denote polar coordinates of $\vec{\mathbf{u}}$ which minimizes $h_1(\vec{\mathbf{u}}|\mathbf{a})$ as $\alpha_{\rm min}$ and $\phi_{\rm min}$, and of $\vec{\mathbf{u}}$ which maximizes $h_1(\vec{\mathbf{u}}|\mathbf{a})$ as $\alpha_{\rm max}$ and $\phi_{\rm max}$. We further denote
	$$
	\alpha_{\rm near}:=\mathop{\arg\min}_{\alpha\in[\alpha_l,\alpha_r]} |\alpha-\alpha_c|,~~\phi_{\rm near}:=\mathop{\arg\min}_{\phi\in[\phi_l,\phi_r]} |\phi-\phi_c|.
	$$ 
	$$
	\alpha_{\rm far}:=\mathop{\arg\max}_{\alpha\in[\alpha_l,\alpha_r]} |\alpha-\alpha_c|,~~\phi_{\rm far}:=\mathop{\arg\max}_{\phi\in[\phi_l,\phi_r]} |\phi-\phi_c|.
	$$
	
	We first give a lemma for $\phi_{\rm max}$ and $\phi_{\rm min}$:
	\begin{lemma}\label{lemma::phi}
		If $(\alpha,\phi)\in\partial\subcubeu$ is an extreme point for $h_1(\vec{\mathbf{u}}|\mathbf{a})$ on the boundaries of cube, one must have
		$\phi_{\rm max}=\phi_{\rm near}$ and $ \phi_{\rm min}=\phi_{\rm far}$.
		
		\begin{proof}
        As the $\phi$ dimension disappears for $\alpha=0$ or $\pi$, we can focus on $\alpha\in(0,\pi)$. The result naturally arises from two observations on the partial derivative~\eqref{eqn::h1_partial_phi}. Firstly, for a fixed $\alpha\in(0,\pi)$, we have $\frac{\partial h_1}{\partial \phi}>0$ if $\phi<\phi_c$, and $\frac{\partial h_1}{\partial \phi}<0$ if $\phi>\phi_c$. Secondly, the partial derivative takes the same absolute value for $\phi_1$ and $\phi_2$ equally distant from $\phi_c$. 
		\end{proof}
	\end{lemma}
	After fixing the value of $\phi$ according to Lemma~\ref{lemma::phi}, we can focus on $\alpha$ by studying~\eqref{eqn::h1_partial_alpha}. 
	\begin{lemma}\label{lemma::alpha}
	Given a fixed value of $\phi$ and for $\alpha_c\not=\pi/2$, the partial derivative~\eqref{eqn::h1_partial_alpha} has a unique zero point $\alpha^*\in[0,\pi]$, and $\alpha^*$ is a \textbf{global maximizer} of $h_1(\vec{\mathbf{u}}|\mathbf{a})$ if $|\phi_c-\phi|<\pi/2$, and a \textbf{global minimizer} of $h_1(\vec{\mathbf{u}}|\mathbf{a})$ if $|\phi_c-\phi|>\pi/2$.
		
		\begin{proof}
			Note that~\eqref{eqn::h1_partial_alpha} can be organized into a form of $A\sin{(\alpha+\bar{\beta})}$, with $\bar{\beta}$ a fixed angle. For $\alpha_c\not=\pi/2$, we have $\bar{\beta}\not=0\text{ or }\pi$, and thus zero point $\alpha^*$ is unique within $[0,\pi]$. 
            
            Given $\alpha_c\not=\pi/2$, $\alpha=\pi/2$ is not a zero point of~\eqref{eqn::h1_partial_alpha}, and thus we can safely rewrite it as:
			\begin{equation}\label{eqn::h1_partial_alpha::transform}
				\frac{\partial h_1}{\partial \alpha}= \cos{\alpha_c}\cos{\alpha}(\tan{\alpha_c}\cos{(\phi_c-\phi)}-\tan{\alpha}).
			\end{equation}
		Based on~\eqref{eqn::h1_partial_alpha::transform}, we list in Table~\ref{h1::table} four cases for the zero point $\alpha^*$ which completes the proof. 
			\begin{table}[htbp]
				\centering
				\caption{ Four different cases of $\alpha_c$ and $\Delta \phi$}
				\begin{tabular}{ccccc}
					\hline
					$\alpha_c$& $|\phi_c-\phi|$ & $\alpha^*$  & $\cos \alpha_c\cos \alpha^*$ & Extrema\\
					\midrule
					$<\pi/2$&  $<\pi/2$ &  $\in(0,\alpha_c)$&  $>0$ & max\\ 
					$>\pi/2$&  $<\pi/2$ &  $\in(\alpha_c,\pi)$&  $>0$ &max\\ 
					$<\pi/2$&  $>\pi/2$ &  $\in(\pi-\alpha_c,\pi)$&  $<0$ &min\\ 
					$>\pi/2$&  $>\pi/2$ &  $\in(0,\pi-\alpha_c)$&  $<0$ & min\\ 
					\bottomrule
				\end{tabular}
				\label{h1::table}
			\end{table}
		\end{proof}
	\end{lemma}
    
\begin{figure}[tbp]
		\centering
		\begin{subfigure}[b]{0.22\textwidth}
			\includegraphics[width=\textwidth]{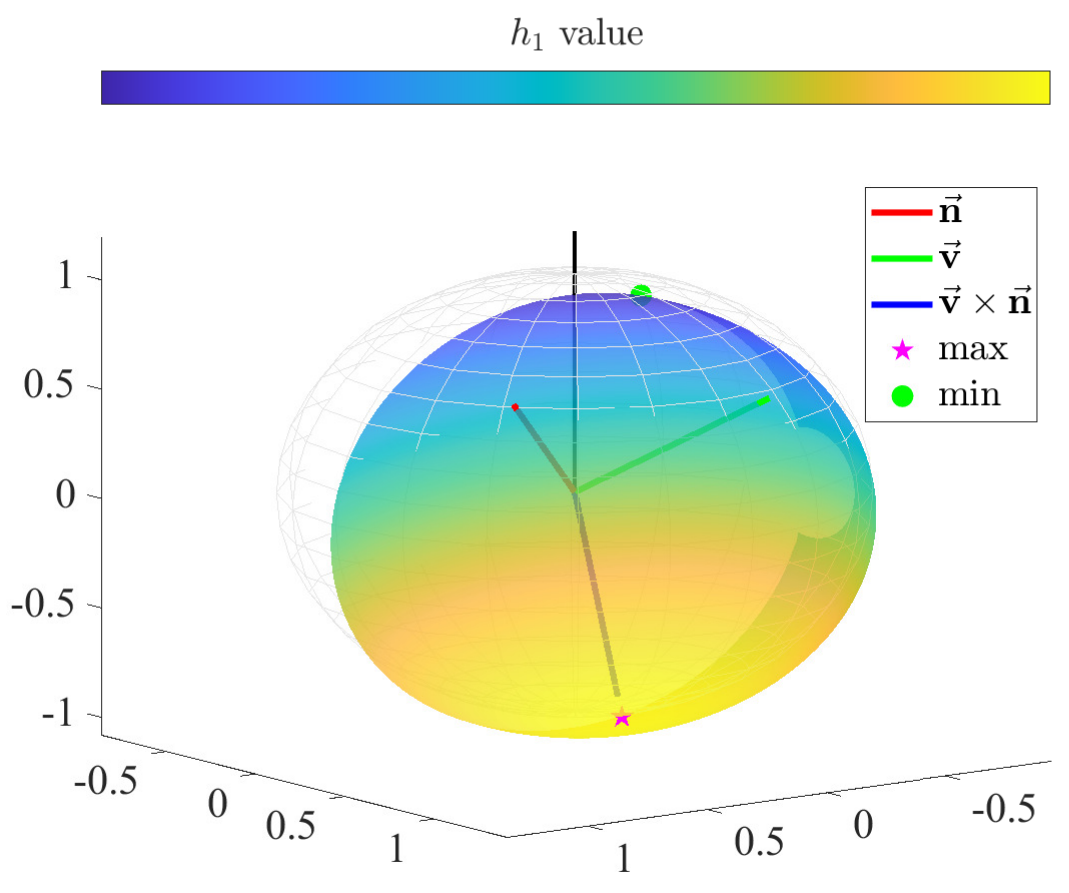}
			\caption{values over a half sphere}
			\label{fig::h1_half_sphere}
		\end{subfigure}
		\hfill
		\begin{subfigure}[b]{0.22\textwidth}
			\includegraphics[width=\textwidth]{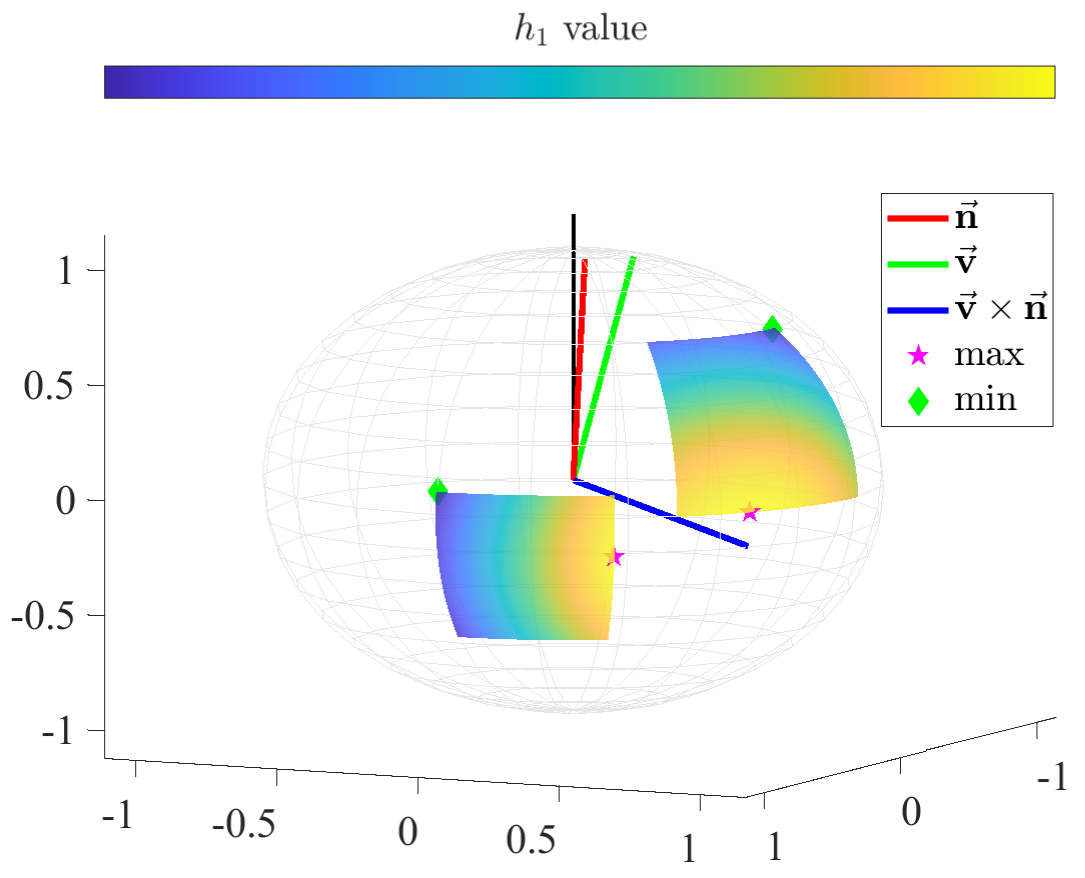}
			\caption{values over a sub-cube}
			\label{fig::h1_sub_cube}
		\end{subfigure}
		\caption{Landscapes of function $h_1(\vec{\mathbf{u}}|\mathbf{a})$.}
		\label{fig::h1_land_scape}
	\end{figure}

Notice that we omit special cases where $\alpha_c=\pi/2$ or $|\phi-\phi_c|=\pi/2$ in Lemma~\ref{lemma::alpha} for conciseness, while interested readers can refer to our implementation for details. Based on Lemmas~\ref{lemma::phi} and~\ref{lemma::alpha}, we propose an efficient procedure to find extreme points of $h_1(\vec{\mathbf{u}}|\mathbf{a})$ on $\partial\subcubeu$ as follows. We introduce the notation $\alpha^*(\delta\phi)$ for the unique zero point of $\alpha$ given $\delta\phi:=|\phi_c-\phi|$, and denote
	$$
	\alpha_{\text{near}}[\alpha^*(\delta\phi)]:=\mathop{\arg\min}_{\alpha\in[\alpha_l,\alpha_r]} |\alpha-\alpha^*(\delta\phi)|,
	$$
	$$
	\alpha_{\text{far}}[\alpha^*(\delta\phi)]:=\mathop{\arg\max}_{\alpha\in[\alpha_l,\alpha_r]} |\alpha-\alpha^*(\delta\phi)|.
	$$
Find the maximizer with $\phi=\phi_{\rm near}$, $\delta\phi_{\rm near}:= |\phi_c-\phi_{\rm near}|$:
	\begin{enumerate}
		\item If $\delta\phi_{\rm near}=0$, we have $\alpha_{\rm max}=\alpha_{\rm near}$;
		
		\item if $\delta\phi_{\rm near}=\pi/2$,
		$$\alpha_{\max}=\begin{cases}
			\alpha_l &\text{if} \quad \alpha_c \le \pi/2, \\
			\alpha_r &\text{if} \quad \alpha_c > \pi/2;
		\end{cases}$$
		\item if $\delta\phi_{\rm near}>\pi/2$, we have $\alpha_{\rm max}=\alpha_{\text{far}}[\alpha^*(\delta\phi_{\rm near})]$
		\item if $\delta\phi_{\rm near}<\pi/2$, $\alpha_c<\pi/2$, and $\alpha_l>=\alpha_c$, we have
		$$
		\alpha_{\rm max}=\alpha_l;
		$$
		\item if $\delta\phi_{\rm near}<\pi/2$, $\alpha_c>\pi/2$ and $\alpha_r<=\pi-\alpha_c$: 
		$$
		\alpha_{\rm max}=\alpha_r;
		$$
		\item otherwise, $\alpha_{\rm max}=\alpha_{\text{near}}[\alpha^*(\delta\phi_{\rm near})]$.
	\end{enumerate}
Find the minimizer with $\phi=\phi_{\rm far}$, $\delta\phi_{\rm far}:=|\phi_c-\phi_{\rm far}|$:
	\begin{enumerate}
		\item If $\delta\phi_{\rm far}<\pi/2$, we have  $\alpha_{\rm min}=\alpha_{\text{far}}[\alpha^*(\delta\phi_{\rm far})]$
		\item if $\delta\phi_{\rm near}=\pi/2$,
		$$
		\alpha_{\min}=\begin{cases}
			\alpha_r &\text{if} \quad \alpha_c \le \pi/2, \\
			\alpha_l &\text{if} \quad \alpha_c > \pi/2;
		\end{cases}
		$$
		\item if $\delta\phi_{\rm far}>\pi/2$, $\alpha_c<\pi/2$ and $\alpha_r<=\pi-\alpha_c$:
		$$
		\alpha_{\rm min}=\alpha_r;
		$$
		\item if $\delta\phi_{\rm far}>\pi/2$, $\alpha_c>\pi/2$ and $\alpha_l>=\pi-\alpha_c$:
		$$
		\alpha_{\rm min}=\alpha_l;
		$$
		\item otherwise, $\alpha_{\rm min}=\alpha_{\text{near}}[\alpha^*(\delta\phi_{\rm near})]$
	\end{enumerate}
	We finally illustrate Theorem~\ref{Theorem::h1} and the above results by visualizing $h_1(\vec{\mathbf{u}}|\mathbf{a})$ on the sphere, as presented in Fig.~\ref{fig::h1_land_scape}.
	\section{Semantic Line Map Construction}\label{appendix::map_construction}
\begin{figure}[!h]
		\centering
		\includegraphics[width=0.85\linewidth]{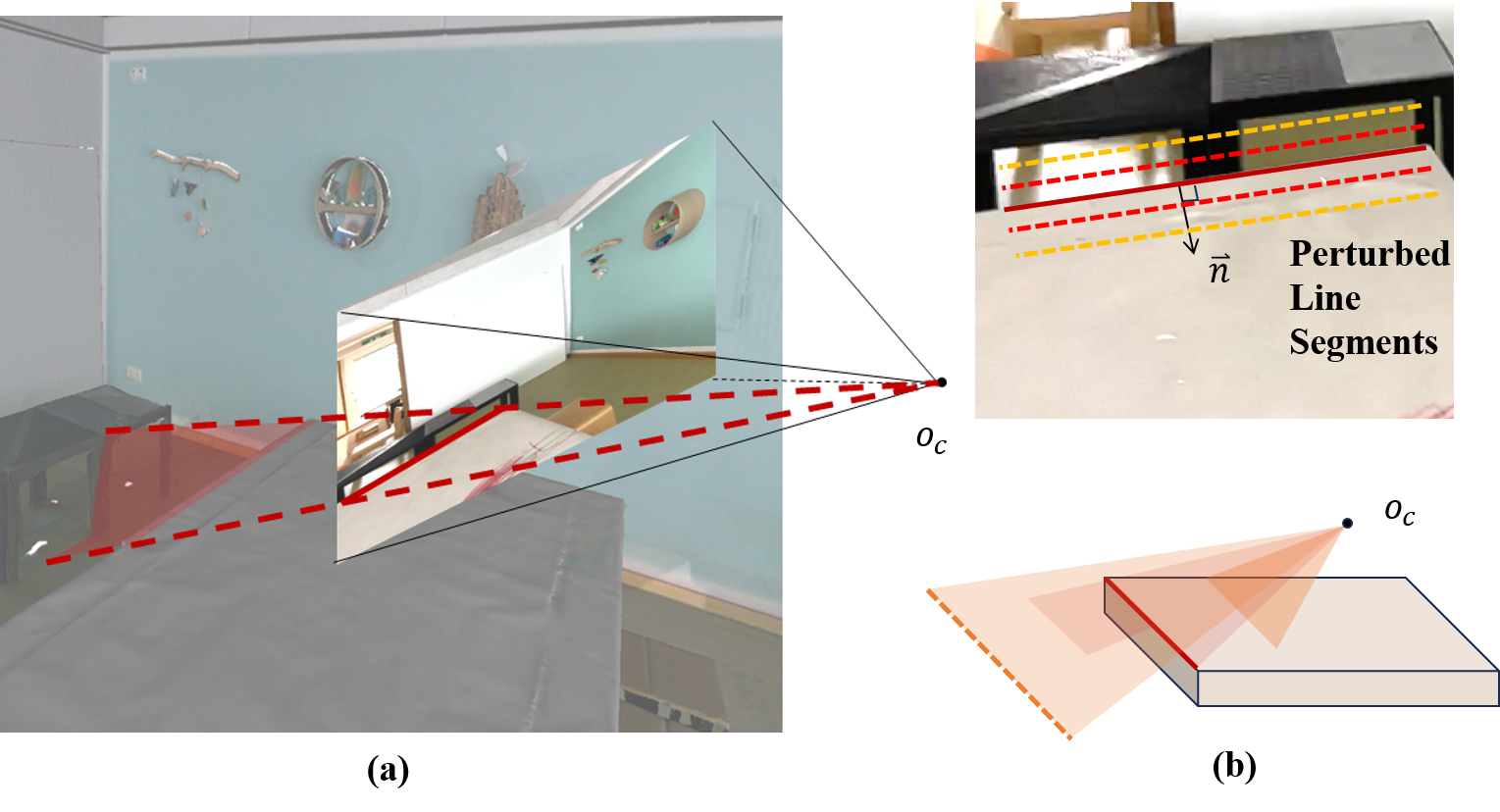}
		\caption{(a) Regressing a 3D line based on points found by inverse projection is prone to the interference of background points. (b) We select the foreground 3D line from multiple hypotheses obtained from perturbed 2D line segments.  
		}
		\label{fig::inverse_projection}
	\end{figure}

	We propose to construct a semantic 3D line map based on posed RGB-D images and their semantic segmentation. The procedure consists of three steps:
	\begin{enumerate}
		\item Extract 2D lines in each image, and assign a semantic label according to segmentation.
		\item Inversely project a 2D line $l$ based on depth and regress a 3D line $L$, assign $L$ with the same semantic as $l$.
		\item After processing all images, cluster and prune the regressed 3D lines.  
	\end{enumerate}
	For the first step, we adopt ELSED~\cite{suarez2022elsed} to extract 2D lines, and use the rendered semantic mask provided by ScanNet++. For the second step, we dedicate to tackling the background interference issue. As shown in Fig.~\ref{fig::inverse_projection}, the inversely projected 3D points can fall on the background while the extracted edge belongs to the foreground object, due to pose or depth error. We propose to evaluate multiple line hypotheses generated from perturbing the 2D line, and select one with a small average depth and mild perturbation. 
	
\begin{figure}
		\centering
		\includegraphics[width=\linewidth]{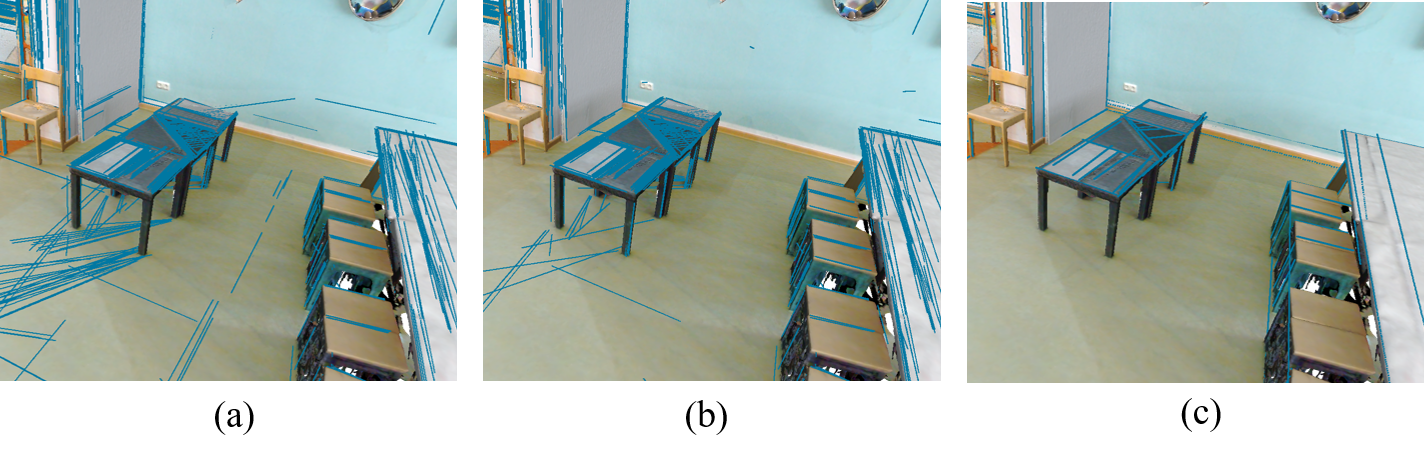}
		\caption{Regressed 3d lines~(a) w/o applying the multi-hypotheses and clustering algorithms,~(b) w/o clustering, and~(c) with both algorithms applied. Note that Thanks to the multi-hypothesis algorithm, there are much less lines in~(b) which fall on the background compared to~(a). The clustering algorithm clusters the proximate lines and prune the less-observed lines in~(b), giving a much neater map in~(c).}
		\label{fig::regression}
	\end{figure}

	Since a 3D line can be observed by multiple images from different viewpoints, there exist a lot of redundant 3D lines after processing all images. In the last step, we cluster these lines based on geometric constraints. We treat each 3D line $L_m:(\mathbf{p}_m,\vec{\mathbf{v}}_m,s_m)$ as a vertex $\mathcal{V}_m$ on a graph $\mathcal{G}$. For two lines $L_m,L_n$ satisfying the following \textbf{parallel} and \textbf{proximity} conditions, we connect the two vertices with an edge: 
	$$
	\angle(\vec{\mathbf{v}}_m,\vec{\mathbf{v}}_,)<\delta_r \textbf{ and } \|(\mathbf{I}_3-\vec{\mathbf{v}}_m\vec{\mathbf{v}}_m^\top)(\mathbf{p}_m-\mathbf{p}_,)\|<\delta_t.$$
	Denote the set of vertex $\mathcal{V}_m$ and its neighbors as $\mathcal{N}(\mathcal{V}_m)$. We summarize the clustering algorithm in Algorithm~\ref{algorithm::clustering}. The overall procedure is fast and efficient with only several parameters to set, which rarely require tuning across different scenes. In fact, we use the same set of parameters across scenes in experiments. As a demonstration of effectiveness, we present local line maps obtained w/ and w/o the proposed multiple hypothesis and clustering algorithms in Fig.~\ref{fig::regression}.

	\begin{algorithm}[!htbp]
		\caption{3D Line Clustering}
		\label{algorithm::clustering}
		\begin{algorithmic}[1]
			\Statex \textbf{Inputs:} graph $\mathcal{G}$, degree threshold $\delta_d$ 
			\Statex $\mathcal{V}_{\rm max}$, $d_{\rm max}=$ \textit{Maximal Degree} $(\mathcal{G})$
			\While{$d_{\rm max}\geq\delta_d$}
			\State Remove $\mathcal{N}(\mathcal{V}_{\rm max})$ from $\mathcal{G}$.  
			\For{each unique label $s_m$ in $\mathcal{N}(\mathcal{V}_{\rm max})$}
			\State Register a 3D line $L:(\vec{\mathbf{v}}_{\rm max},\mathbf{p}_{\max},s_m)$.
			\EndFor
			\EndWhile
		\end{algorithmic}
	\end{algorithm}

	\section{Find Bounding Functions for Translation Problem in PnL}\label{appendix::translation_bounding}
	For the three degrees of freedom $t_x$, $t_y$ and $t_z$ in translation, we distinguish the one with largest range, as determined by the environment size. Without generality, we assume $t_x$ is the distinguished parameter. Denote $\mathbf{R}_\sigma^*\vec{\mathbf{n}}_a$ as $\vec{\mathbf{n}}_a^*:=(n_{a,x}^*,n_{a,y}^*,n_{a,z}^*)$, and denote $\text{const}_{a}:=\vec{\mathbf{n}}_a^*\bullet\mathbf{p}_{a}$, we rewrite the residual function $f(t_x,t_y,t_z|\mathbf{a})$ in~\eqref{model::satCM_trans} as follows:
	\begin{equation}\label{eqn::trans_residual}
		t_xn_{a,x}^*+t_y n_{a,y}^*+t_zn_{a,z}^*-\text{const}_{a}.
	\end{equation}
	For $t_x$ in its range $\mathcal{I}_x$ and $(t_y,t_z)$ in a sub-cube $\boldsymbol{\mathcal{C}}_{yz}$, the bounding functions $f_L(t_x)$ and $f_U(t_x)$ for~\eqref{eqn::trans_residual} can be easily found by solving a linear programming problem about $(t_y,t_z)$. Take $f_L(t_x)$ as an example. We obtain it by solving
	$$
	\arg\min_{t_y,t_z}~t_yn_{a,y}^*+t_zn_{a,z}^*\quad(t_y,t_z)\in \boldsymbol{\mathcal{C}}_{yz},
	$$
	and substituting the solution into~\eqref{eqn::trans_residual}. Since a global optimum of linear programming occurs at the vertices, we simply compare the value achieved by the four vertices of $\boldsymbol{\mathcal{C}}_{yz}$. 

	\section{Supplementary Experiment Results}\label{appendix::supplementray}
	\subsection{Sat-CM v.s. CM in translation estimation estimation}
	We compare translation estimation accuracy between the Sat-CM and CM method. Since the translation problem~\eqref{model::satCM_trans} is formulated based on a rotation estimate, the orientation accuracy directly governs the precision of translation. In this experiment, we use the most accurate upstream rotation estimates, i.e., those based on the likelihood saturation function~\eqref{eqn::likelihood_sat_func}, to formulate the translation problem. In the legend we use `$\pi-$' and `$\pi/2-$' to distinguish the upstream rotation estimates. We set the translation residual upper bound~(used in Assumption~\ref{assump::uniform}) at $u_t=1$, and choose $q=0.9$ for \textcolor{blue}{SCM}-GT, $q=0.5$ for \textcolor{blue}{SCM}-PR. After solving~\eqref{model::satCM_trans}, we improve the output translation candidates with two steps. Firstly, we prune the inlier associations by imposing two physical constraints: the truly-associated 3D line resides in front of the camera and its projection intersects with the image. Next, we fine-tune the translation estimates by minimizing least squares error or the pruned inliers. As present in Table~\ref{table::trans_SCM_CM}, Sat-CM performs consistently better than CM across all scenes and settings. Notably, the truncated saturation function performs as good as the likelihood saturation function when using ground truth labels, and presents better robustness under predicted labels by keeping more potentially good candidates for further refinement. 
	\begin{table*}
		\centering
		\caption{Compare CM with Sat-CM in translation.}
		\label{table::trans_SCM_CM}
		\begin{tabular}{ccccccccc} 
			\toprule
			\multirow{2}{*}{Method} & S1(workstation) &  S2(office) & S3(game bar) & S4(art room)           & S1  &  S2 & S3 & S4 \\
			& \multicolumn{4}{c}{25\%/50\%/75\% error quantiles~(cm)}         & \multicolumn{4}{c}{Recall at 5cm/10cm/15cm~(\%)}\\
			\midrule
			$\pi$-CM-GT        &3.0/5.7/33.9 & 2.8/4.0/7.9 & 4.6/9.9/78.4 & 3.3/5.5/34.9 & 47/65/72 & 60/78/81 & 25/50/61 & 42/68/71\\
			$\pi$-SCM0-GT      &2.9/5.9/11.0 & 2.4/3.8/7.1 & 4.0/7.1/22.2 & 2.9/5.0/8.5 & 45/73/80 & 63/84/88 & 30/58/68 & 50/77/84\\
			$\pi$-{\color{blue}SCM1}-GT &3.0/5.8/14.1 & 2.5/4.0/6.7 & 4.0/6.8/22.6 & 3.0/5.0/8.2 & 44/71/76 & 66/84/88 & 30/58/68 & 50/82/85\\
			
			\midrule
			$\pi/2$-CM-GT                &3.0/5.5/15.6 & 2.8/4.0/7.9 & 4.2/8.6/28.1 & 3.3/5.5/19.7 & 48/66/75 & 60/78/82 & 26/51/63 & 42/68/71\\
			$\pi/2$-SCM0-GT               &2.8/5.9/10.8 & 2.4/3.8/6.7 & 4.0/6.9/19.4 & 2.9/4.9/8.2 & 46/74/85 & 64/85/91 & 32/61/72 & 51/78/86\\
			$\pi/2$-{\color{blue}SCM1}-GT &2.7/5.6/11.3 & 2.5/3.8/6.6 & 4.0/6.5/18.9 & 3.0/4.9/8.0 & 45/72/80 & 67/85/91 & 32/62/72 & 51/82/88\\
			
			\midrule
			$\pi$-CM-PR   &5.2/77.4/198.4 & 5.2/63.0/214.9 & 8.2/129.1/241.2 & 6.9/121.5/226.9 & 24/37/42 & 23/41/44 & 17/27/35 & 18/29/34\\
			$\pi$-SCM0-PR  &4.8/13.3/150.7 & 3.7/7.7/197.5 & 7.2/100.7/232.0 & 5.9/21.2/195.5 & 28/44/53 & 35/59/62 & 17/29/37 & 20/39/45\\
			$\pi$-{\color{blue}SCM1}-PR &4.5/13.8/150.7 & 3.7/7.3/194.5 & 6.4/135.8/226.6 & 5.9/33.8/208.3 & 28/44/53 & 33/58/61 & 19/33/41 & 20/38/45\\
			
			\midrule
			$\pi/2$-CM-PR   &5.2/16.7/175.2 & 4.7/11.8/194.5 & 7.7/61.1/234.9 & 6.4/107.0/208.1 & 24/42/48 & 26/47/52 & 17/29/37 & 19/31/36\\
			$\pi/2$-SCM0-PR  &4.4/10.6/90.1 & 3.7/6.8/49.2 & 6.6/32.3/199.0 & 5.5/14.4/119.8 & 30/49/60 & 36/65/70 & 19/33/43 & 23/43/51\\                     
			$\pi/2$-{\color{blue}SCM1}-PR &4.5/10.2/113.2 & 3.6/6.3/105.9 & 5.2/26.7/224.5 & 5.7/18.0/192.1 & 29/49/59 & 36/65/70 & 24/40/49 & 23/41/48\\
			
			\bottomrule
		\end{tabular}
	\end{table*}

	\subsection{parameter sensitivity evaluation}
\begin{figure}[!b]
		\centering
		\begin{subfigure}[b]{0.22\textwidth}
			\includegraphics[width=\textwidth]{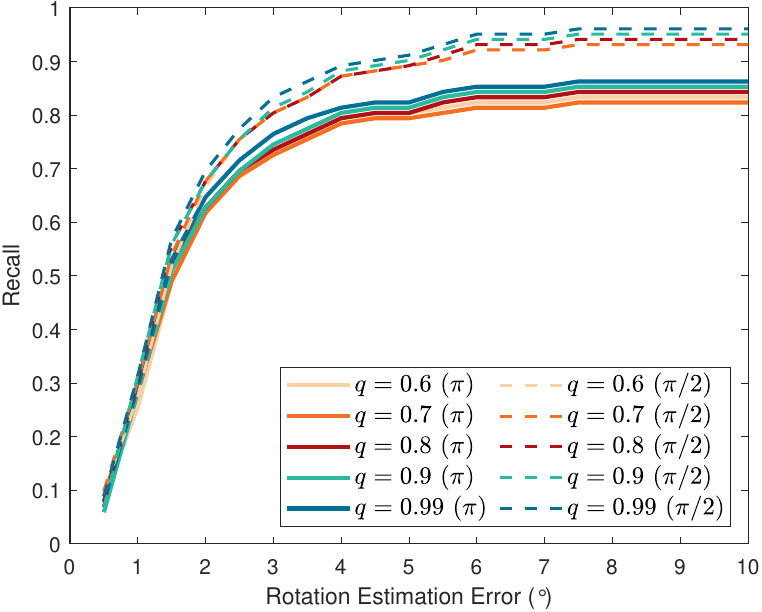}
			\caption{S1~work station-GT}
		\end{subfigure}
		\hfill
		\begin{subfigure}[b]{0.22\textwidth}
			\includegraphics[width=\textwidth]{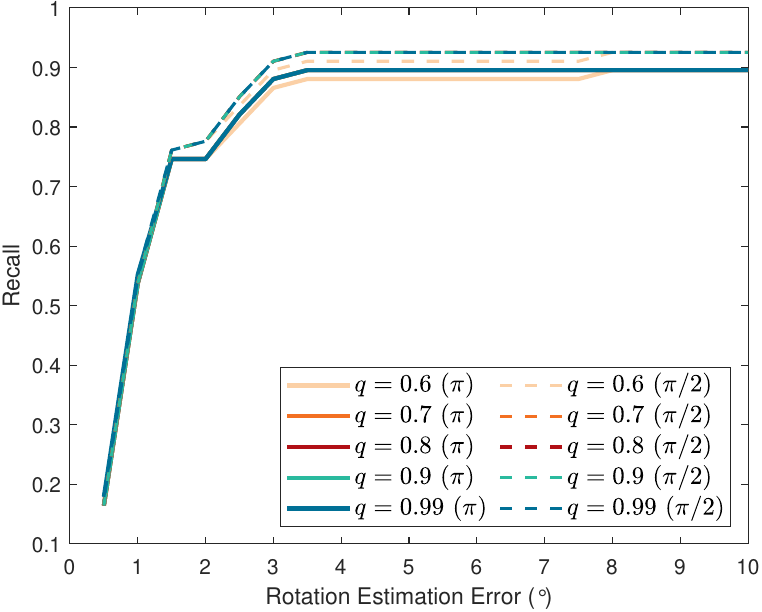}
			\caption{S2~office-GT}
		\end{subfigure}
		\hfill
		\begin{subfigure}[b]{0.22\textwidth}
			\includegraphics[width=\textwidth]{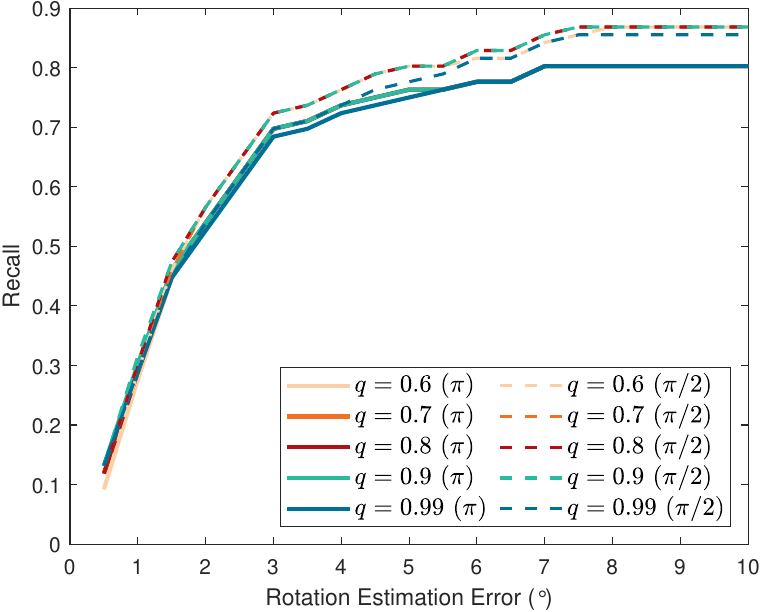}
			\caption{S3~game bar-GT}
		\end{subfigure}
		\hfill
		\begin{subfigure}[b]{0.22\textwidth}
			\includegraphics[width=\textwidth]{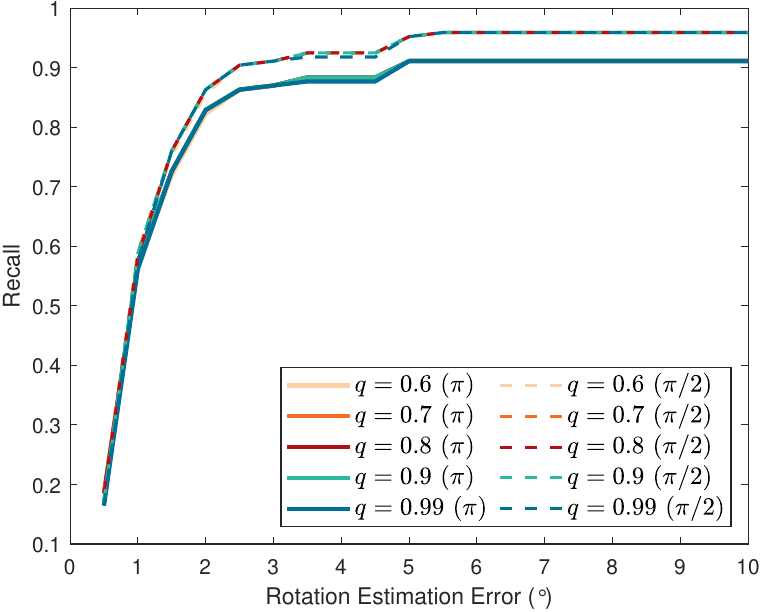}
			\caption{S4~art room-GT}
		\end{subfigure}
		\caption{Rotation error recall using ground truth semantic labels and different values of $q$.}
		\label{fig::sensitivity_gt}
	\end{figure}

\begin{figure*}
		\centering
		\begin{subfigure}[b]{0.22\textwidth}
			\includegraphics[width=\textwidth]{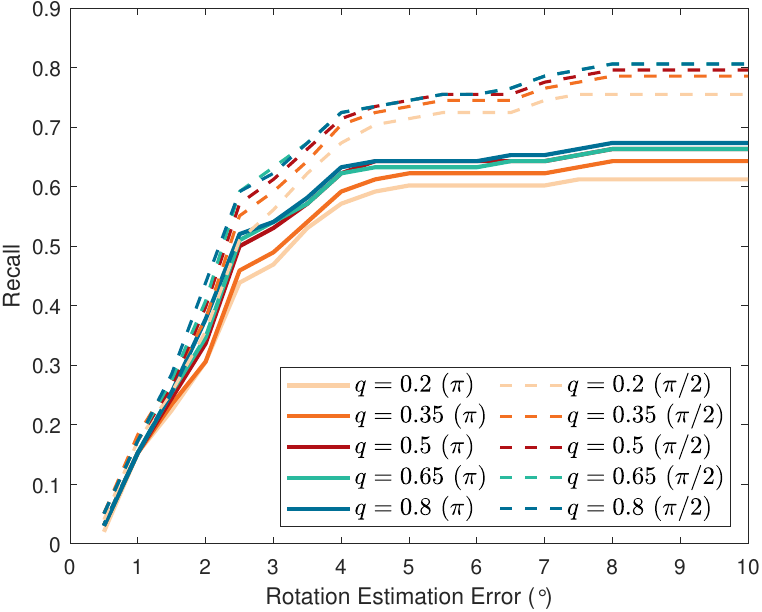}
			\caption{S1~work station-PR}
		\end{subfigure}
		\hfill
		\begin{subfigure}[b]{0.22\textwidth}
			\includegraphics[width=\textwidth]{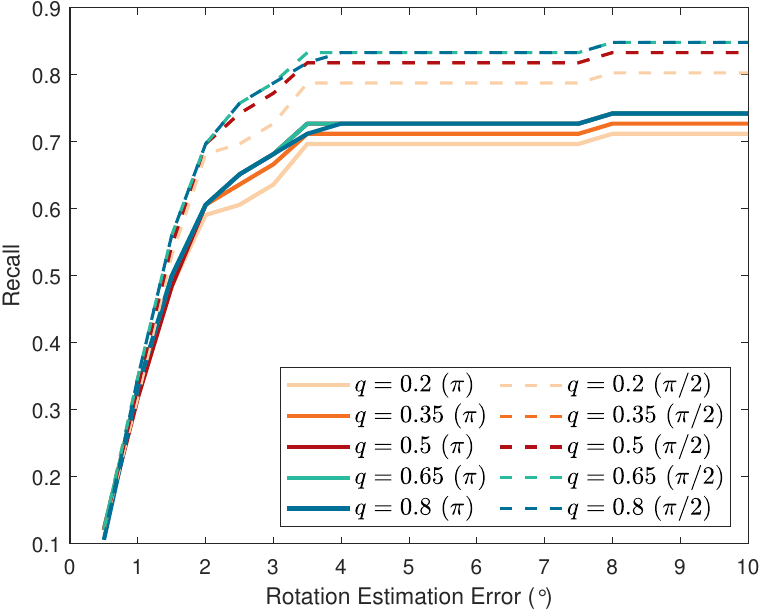}
			\caption{S2~office-PR}
		\end{subfigure}
		\hfill
		\begin{subfigure}[b]{0.22\textwidth}
			\includegraphics[width=\textwidth]{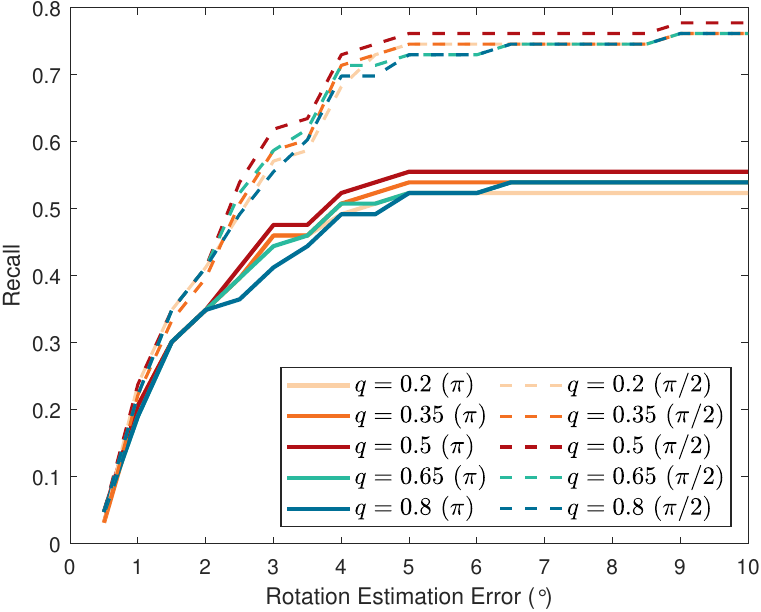}
			\caption{S3~game bar-PR}
		\end{subfigure}
		\hfill
		\begin{subfigure}[b]{0.22\textwidth}
			\includegraphics[width=\textwidth]{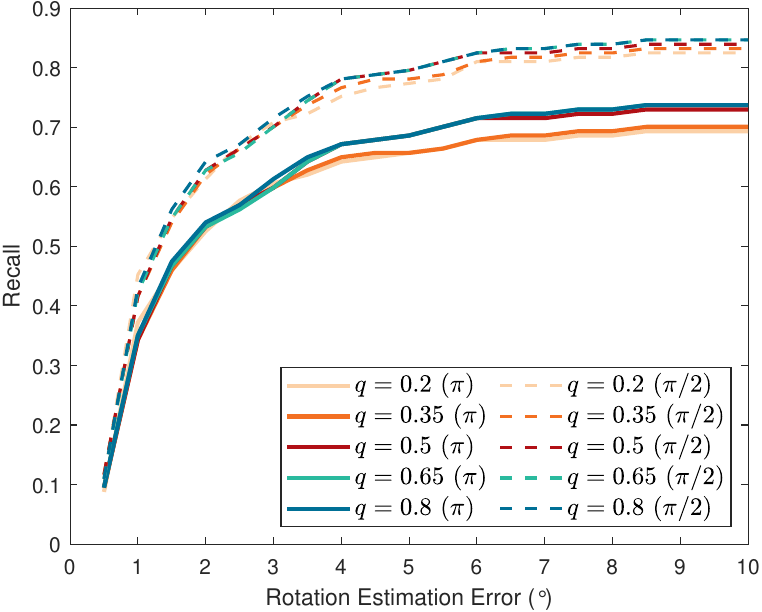}
			\caption{S4~art room-PR}
		\end{subfigure}
		\caption{Rotation error recall using predicted semantic labels and different values of $q$.}
		\label{fig::sensitivity_pr}
	\end{figure*}

\begin{figure*}
		\centering
		\begin{subfigure}[b]{0.22\textwidth}
			\includegraphics[width=\textwidth]{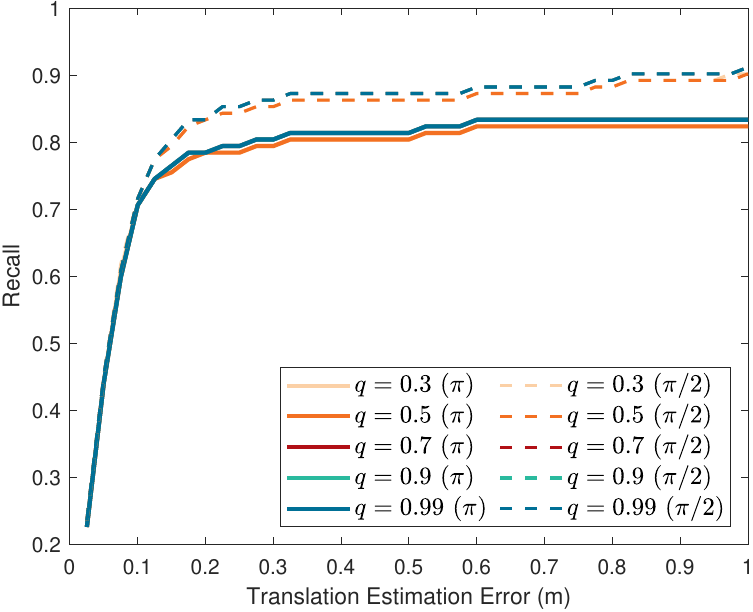}
			\caption{S1~work station-GT}
		\end{subfigure}
		\hfill
		\begin{subfigure}[b]{0.22\textwidth}
			\includegraphics[width=\textwidth]{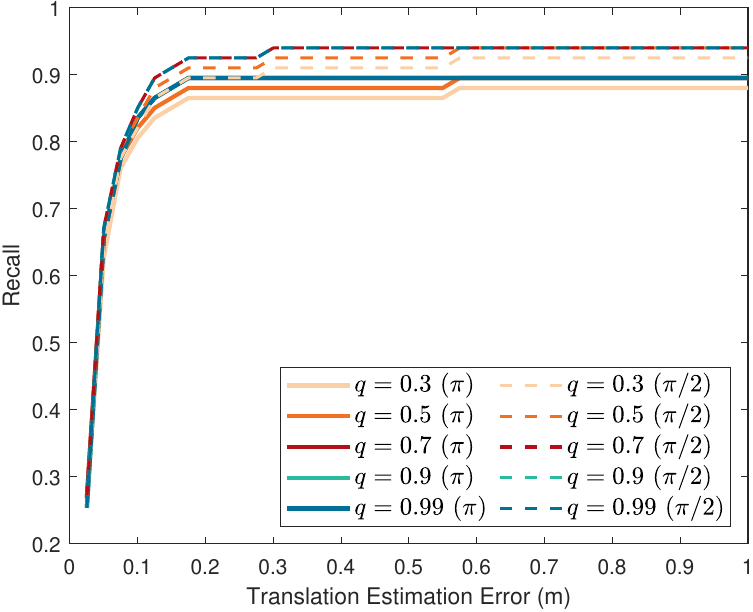}
			\caption{S2~office-GT}
		\end{subfigure}
		\hfill
		\begin{subfigure}[b]{0.22\textwidth}
			\includegraphics[width=\textwidth]{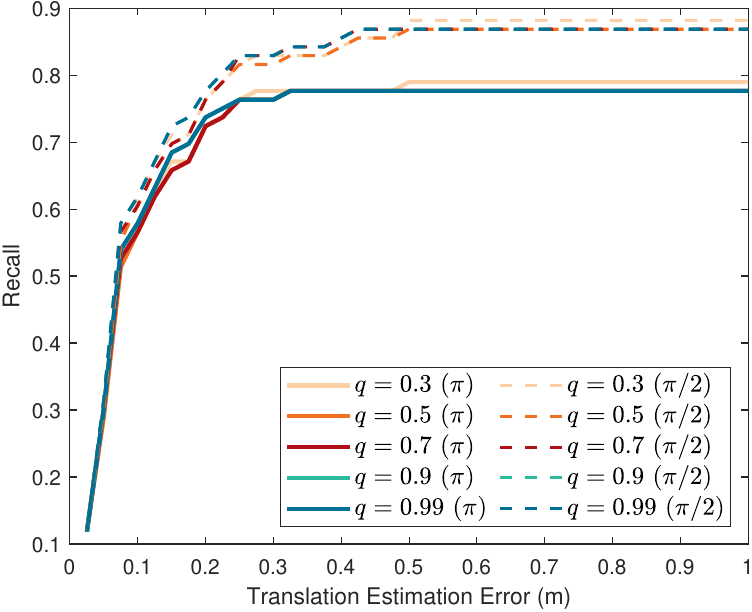}
			\caption{S3~game bar-GT}
		\end{subfigure}
		\hfill
		\begin{subfigure}[b]{0.22\textwidth}
			\includegraphics[width=\textwidth]{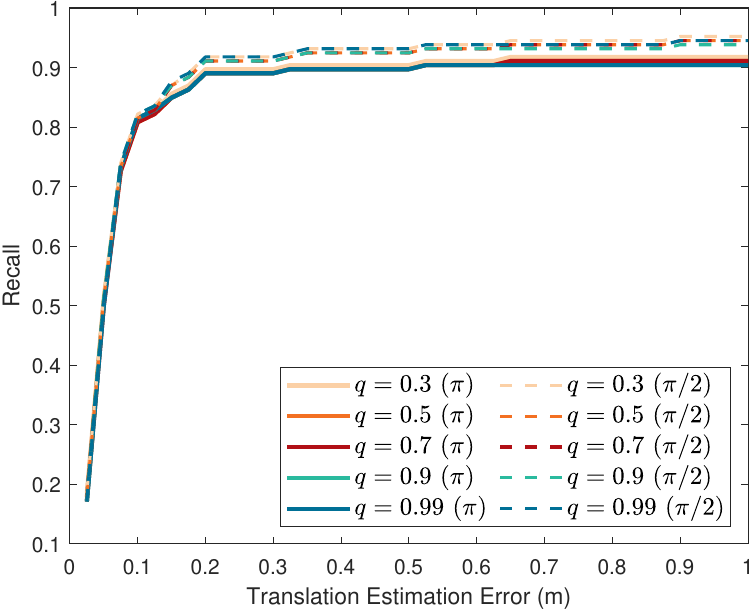}
			\caption{S4~art room-GT}
		\end{subfigure}
		\caption{Translation error recall using ground truth semantic labels and different values of $q$.}
		\label{fig::sensitivity_trans_gt}
	\end{figure*}

\begin{figure*}
		\centering
		\begin{subfigure}[b]{0.22\textwidth}
			\includegraphics[width=\textwidth]{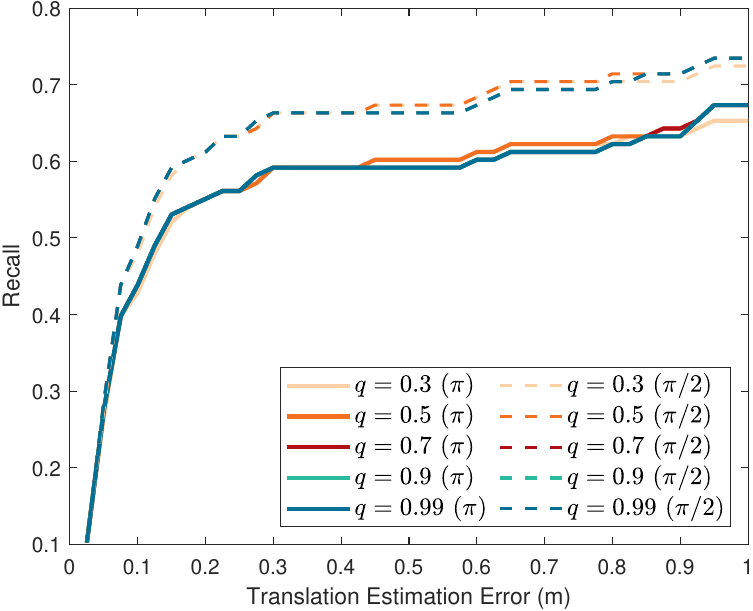}
			\caption{S1~work station-PR}
		\end{subfigure}
		\hfill
		\begin{subfigure}[b]{0.22\textwidth}
			\includegraphics[width=\textwidth]{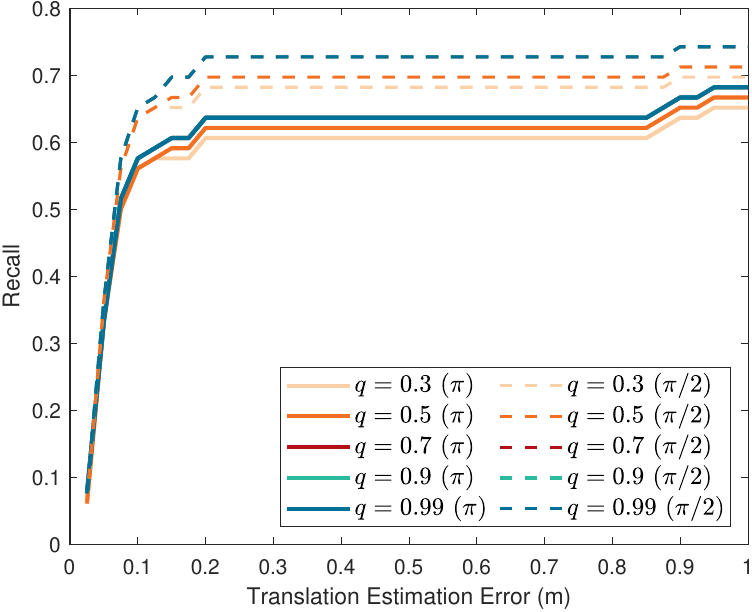}
			\caption{S2~office-PR}
		\end{subfigure}
		\hfill
		\begin{subfigure}[b]{0.22\textwidth}
			\includegraphics[width=\textwidth]{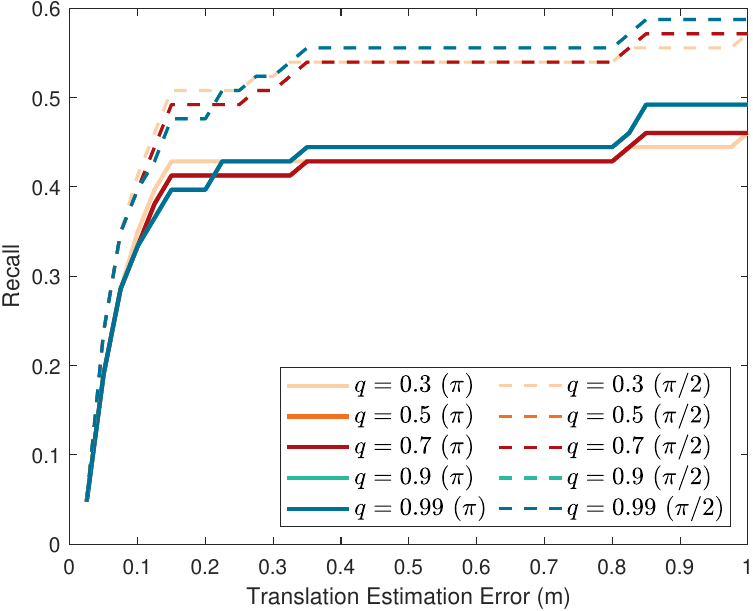}
			\caption{S3~game bar-PR}
		\end{subfigure}
		\hfill
		\begin{subfigure}[b]{0.22\textwidth}
			\includegraphics[width=\textwidth]{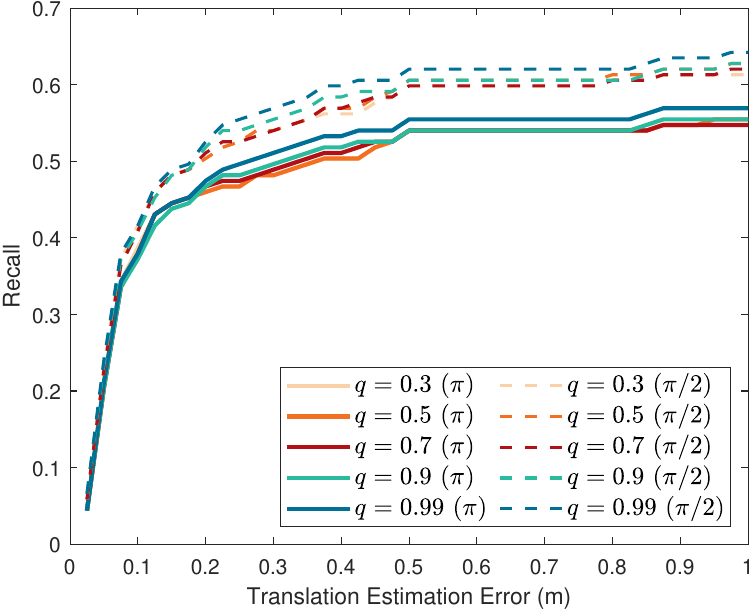}
			\caption{S4~art room-PR}
		\end{subfigure}
		\caption{Translation error recall using predicted labels and different values of $q$.}
		\label{fig::sensitivity_trans_pr}
	\end{figure*}

	We evaluate performance sensitivity in parameter $q$ by plotting the error curves. For rotation estimation, we choose 
	$
	q\in\{0.6,0.7,0.8,0.9,0.99\}
	$
	when using the ground truth labels and choose 
	$
	q\in\{0.2,0.35,0.5,0.65,0.8\}
	$
	when using the predicted labels. The results are present in Fig.~\ref{fig::sensitivity_gt} and~\ref{fig::sensitivity_pr}. For translation estimation, we choose the same set of $q$ values for both ground truth and predicted semantic labels $
	q\in\{0.1,0.3,0.5,0.7,0.9\}. 
	$ The results are present in Fig.~\ref{fig::sensitivity_trans_gt} and Fig.~\ref{fig::sensitivity_trans_pr}.

\end{document}